\documentclass{article}

\usepackage{microtype}
\usepackage{graphicx}
\usepackage{caption}
\usepackage{subcaption}
\usepackage{booktabs} 
\usepackage{hyperref}
\usepackage{colortbl}
\usepackage{url}
\usepackage{makecell}
\usepackage{stfloats}
\usepackage[dvipsnames]{xcolor}
\usepackage{soul}
\usepackage{ulem}
\usepackage{listings}
\usepackage[numbers]{natbib}
\usepackage{enumitem}
\usepackage{wrapfig2}

\usepackage[utf8]{inputenc}
\usepackage[english]{babel}
\usepackage{times}
\usepackage{indentfirst}
\usepackage[dvipsnames]{xcolor}
\definecolor{codegreen}{rgb}{0,0.6,0}
\definecolor{codegray}{rgb}{0.5,0.5,0.5}
\definecolor{codepurple}{rgb}{0.58,0,0.82}
\definecolor{backcolour}{rgb}{0.95,0.95,0.92}
\lstdefinestyle{mystyle}{
    backgroundcolor=\color{backcolour},   
    commentstyle=\color{codepurple},
    keywordstyle=\color{NavyBlue},
    numberstyle=\tiny\color{codegray},
    stringstyle=\color{codepurple},
    basicstyle=\ttfamily\footnotesize\bfseries,
    breakatwhitespace=false,         
    breaklines=true,                 
    captionpos=t,                    
    keepspaces=true,                 
    numbers=left,                    
    numbersep=5pt,                  
    showspaces=false,                
    showstringspaces=false,
    showtabs=false,                  
    tabsize=2
}
\usepackage[acronym]{glossaries}
\usepackage{amsmath, amssymb, dsfont, mathtools, amsthm, amsfonts, nicefrac}
\usepackage{float}
\usepackage[title]{appendix}
\usepackage{multirow}
\usepackage{pifont}
\usepackage{times}
\usepackage[capitalise, noabbrev]{cleveref}
\usepackage[nonatbib, final]{neurips_2025}

\theoremstyle{plain}
\newtheorem{theorem}{Theorem}[section]
\newtheorem{proposition}[theorem]{Proposition}

\theoremstyle{definition}

\theoremstyle{remark}

\DeclareMathOperator*{\argmax}{arg\,max}

\newcommand{\meanstd}[2]{#1\tiny{\(\pm\)#2}}

\newcommand{\mysection}[2][]{
    \ifthenelse{ \equal{#1}{} }
    {\section{\texorpdfstring{#2}{#2}}}
    {\section{\texorpdfstring{#2}{#1}}}
}
\newcommand{\mysubsection}[2][]{
    \ifthenelse{ \equal{#1}{} }
    {\subsection{\texorpdfstring{#2}{#2}}}
    {\subsection{\texorpdfstring{#2}{#1}}}
}
\newcommand{\mysubsubsection}[2][]{
    \ifthenelse{ \equal{#1}{} }
    {\subsubsection{\texorpdfstring{#2}{#2}}}
    {\subsubsection{\texorpdfstring{#2}{#1}}}
}

\hypersetup{
  colorlinks=true,
  citecolor=MidnightBlue,
  linkcolor=RoyalPurple,
  final
}

\newglossaryentry{true_p}
{
    name={balanced MTPP prediction problem},
    description={The main problem we want to tackle in this paper.}
}

\newglossaryentry{main_p}
{
    name={mark-oriented next event prediction problem},
    description={The main problem we want to tackle in this paper.}
}

\newglossaryentry{c_main_p}
{
    name={time-event prediction problem},
    description={The main problem that been solved in multiple ways.}
}

\newacronym{tpp}{TPP}{Temporal Point Process}
\newacronym{stpp}{STPP}{Spatio-temporal Point Process}
\newacronym{mtpp}{MTPP}{Marked Temporal Point Process}
\newacronym{pdf}{PDF}{Probability Distribution Function}
\newacronym{cdf}{CDF}{Cumulative Distribution Function}
\newacronym{ddpm}{DDPM}{Denoising Diffusion Probabilistic Model}
\newacronym{cif}{CIF}{Conditional Intensity Function}
\newacronym{model}{IFNMTPP}{Integration-free Neural Marked Temporal Point Process}
\newacronym{iem}{IEM}{Integral Estimation Module}
\newacronym{ta}{TA}{Thinning Algorithm}
\newacronym{te}{NEP}{Next-event Prediction problem}
\newacronym{et}{MT paradigm}{Mark-time prediction paradigm}
\newacronym{ewmae}{EW-MAE}{event-oriented MAE}
\newacronym{its}{ITS}{Inverse Transform Sampling}
\newacronym{mmae}{MAE}{Mean Absolute Error}
\newacronym{mnep}{RM-NEP}{Rare-event-aware Next Event Prediction}

\title{Addressing Mark Imbalance in Integration-free Neural Marked Temporal Point Processes}

\author{
  Sishun Liu \\
  RMIT University \\
  Melbourne, Victoria 3000 \\
  \texttt{sishun.liu@student.rmit.edu.au} \\
  \And 
  Ke Deng \\
  RMIT University \\
  Melbourne, Victoria 3000 \\
  \texttt{ke.deng@rmit.edu.au} \\
  \And
  Yongli Ren \\
  RMIT University \\
  Melbourne, Victoria 3000 \\
  \texttt{yongli.ren@rmit.edu.au} \\
  \And
  Yan Wang \\
  Macquarie University \\
  Syndey, New South Wales 2000 \\
  \texttt{yan.wang@mq.edu.au} \\
  \And 
  Xiuzhen Zhang \\
  RMIT University \\
  Melbourne, Victoria 3000 \\
  \texttt{xiuzhen.zhang@rmit.edu.au} \\
}

\begin{document}
\maketitle

\begin{abstract}
\acrfull*{mtpp} has been well studied to model the event distribution in marked event streams, which can be used to predict the mark and arrival time of the next event. However, existing studies overlook that the distribution of event marks is highly imbalanced in many real-world applications, with some marks being frequent but others rare. The imbalance poses a significant challenge to the performance of the next event prediction, especially for events of rare marks. To address this issue, we propose a \textit{thresholding} method, which learns thresholds to tune the mark probability normalized by the mark's prior probability to optimize mark prediction, rather than predicting the mark directly based on the mark probability as in existing studies. In conjunction with this method, we predict the mark first and then the time. In particular, we develop a novel neural \acrshort*{mtpp} model to support effective time sampling and estimation of mark probability without computationally expensive numerical improper integration. Extensive experiments on real-world datasets demonstrate the superior performance of our solution against various baselines for the next event mark and time prediction. The code is available at \url{https://github.com/undes1red/IFNMTPP}.
\end{abstract}

\section{Introduction}\label{sec:introduction}
\textit{\acrfull*{mtpp}} models event sequences observed from natural phenomena (e.g. earthquakes) or generated in human activities (e.g. retweets), where each event has a mark and an arrival time. \acrshort*{mtpp} has attracted the attention of the research community (see \citet{Shchur2021} for a comprehensive review). Typically, \acrshort*{mtpp} models the joint \textit{\acrfull*{pdf}} conditioned on history, denoted as \(p^*(m, t)\)\footnote{The asterisk reminds the probability is conditioned on history, i.e., the events in the past.}, where \(m\) and \(t\) are the mark and arrival time\footnote{The time relative to the most recent event} of the next event, respectively. Some studies are on the Poisson Process \citep{xiaoXTSFormerCrossTemporalScaleTransformer2024} and Hawkes process \citep{rizoiu_expecting_2017,ide_cardinality-regularized_2021,okawa_dynamic_2021}. Recently, we have witnessed a rapid growth of \textit{neural \acrshort*{mtpp}}, which models \(p^*(m, t)\) using neural networks \citep{mei_neural_2017, omi_fully_2019, zhang_self-attentive_2020, mei_transformer_2021, zhouAutomaticIntegrationSpatiotemporal2023}, due to the capability of learning complicated temporal patterns and computational efficiency \citep{Shchur2021}.

However, existing studies overlook that the distribution of event marks is highly imbalanced in many real-world applications, with some marks being frequent but others rare, as shown in \cref{fig:1} (a). Similar to other machine learning tasks such as classification, the imbalance poses a significant challenge to the performance of the next event prediction, especially for events of rare marks, which are often more important than other marks (e.g., the occurrence of a 7-magnitude earthquake or a retweet from celebrities). By mitigating the impact of mark imbalance, this study aims to improve the performance of \acrshort*{mtpp} for next event prediction.

Various techniques have been investigated to improve the prediction performance of rare classes in classifiers, including \textit{resampling the training set}, \textit{cost-sensitive approaches}, and \textit{thresholding} \citep{johnsonSurveyDeepLearning2019, aguiarSurveyLearningImbalanced2024, yang_survey_2022}. Training data resampling requires a proper resampling ratio. Cost-sensitive approaches require domain knowledge on the importance of different marks in setting the cost \citep{johnsonSurveyDeepLearning2019}. To have a solution with minimum external knowledge and assumptions, this study adopts thresholding, which learns thresholds to tune the mark probability normalized by the prior probability of marks.


Addressing mark imbalance for \acrshort*{mtpp} using thresholding is not straightforward. In addition to mark prediction, \acrshort*{mtpp} also needs to predict the time 
simultaneously. In most existing \acrshort*{mtpp} studies, the strategy is to predict the time based on $p^*(t)$, the probability that the next event time is $t$, and then predict the mark based on $p^*(m|t)$, the probability that the next event mark is $m$ at the predicted time $t$. Our analysis and experiments show that this strategy is unsuitable for addressing mark imbalance with thresholding. If time changes
, the mark probability conditioned on time typically changes and thus requires different tuning thresholds. However, it is implausible to learn the tuning thresholds at all times. So, we propose a strategy that first predicts the mark based on $p^*(m)$, the probability that the next event mark is $m$, and then predicts the time based on $p^*(t|m)$, the probability that the next event time is $t$ on the condition that the predicted mark $m$ is the next event mark. Since the mark probability $p^*(m)$ is independent of time, applying thresholding to handle mark imbalance is easy. 

However, our strategy has its challenges. First, two different improper integrations are required for modeling $p^*(m)$ and time prediction, respectively. Second, sampling $p^*(t|m)$ to predict time is inefficient because it needs the \textit{Cumulative Distribution Function (CDF)} of $p^*(t|m)$, but the CDF does not have a closed-form expression. To overcome these challenges, we find a way to unify the two improper integrations into one. Then, we develop a novel \acrshort*{mtpp} model, called \textit{\acrfull*{model}}, to approximate the unified improper integration, rather than using a computationally expensive numerical method. With \acrshort*{model}, we can directly model $p^*(m)$ and the CDF of $p^*(t|m)$. The CDF makes drawing samples from $p^*(t|m)$ efficient for time prediction. Based on $p^*(m)$, the thresholding method can be applied to address the mark imbalance. Extensive experiments on real-world datasets demonstrate the superior performance of our solution against various baselines for the next event mark and time prediction. The contributions of this study are threefold: 

\begin{itemize}[itemsep=0em]
    \item This study investigates the impact of mark imbalance on \acrshort*{mtpp} for next event prediction, which is overlooked by existing \acrshort*{mtpp} studies.
    \item This study introduces the first solution to address mark imbalance in \acrshort*{mtpp}, which learns thresholds to tune the mark probability normalized by the prior probability of marks to optimize mark prediction, rather than predicting marks based on mark probability directly as in existing studies.
    \item This study finds a way to unify two improper integrations into one, and proposes a novel \acrfull*{model} to approximate the unified improper integration to support time sampling and estimation of mark probability, rather than using computationally expensive numerical improper integration.
\end{itemize}

\section{Preliminaries}\label{sec:problem}
\subsection{Marked Temporal Point Process}\label{sec:pre} 
The \acrfull*{mtpp} is a random process whose embodiment is a sequence of discrete events, $\mathcal{S} = \{(m_i,t_i)\}_{i=1}$, where $i\in \mathbb{Z}^+$ is the sequence order, $t_i\in \mathbb{R}^+$ is the time when the $i$th event occurs, $m_i$ is the mark of the $i$th event. This study only concerns a finite set of categorical marks $\mathrm{M}=\{k_1,k_2,\cdots,k_{|\mathrm{M}|}\}$, and the simple \acrshort*{mtpp}, which allows at most one event at every time, thus $t_i<t_j$ if $i<j$. The time of the most recent event is $t_l$, and the current time is $t>t_l$. The time interval between two adjacent events is the inter-event time. We assume that an event with a particular mark at a particular time may be triggered by past events. Let \(\mathcal{H}_{t_l}\) be the history up to (including) the most recent event, and \(\mathcal{H}_{t-}\) be the history up to (excluding) the current time \citep{rasmussen_lecture_2018}. With these definitions, we can define the \textit{\acrfull*{cif}} of \acrshort*{mtpp}:
\begin{equation}
    \lambda^*(m = k_i,t) = \lambda(m = k_i,t|\mathcal{H}_{t-}) =  \lim_{\Delta t \rightarrow 0}\frac{P(m = k_i, t \in [t, t+\Delta t)|\mathcal{H}_{t-})}{\Delta t}.
    \label{eqn:1}
\end{equation} 
With $\lambda^*(m, t)$, the conditional joint \acrshort*{pdf} of the next event can be defined:

\begin{equation}
p^*(m, t) = p(m,t|\mathcal{H}_{t_l}) = \lambda^*(m, t) F^*(t) =  \lambda^*(m, t)\exp({-\int_{t_{l}}^t{\sum_{n\in \mathrm{M}}\lambda^*(n, \tau)d\tau}}).
\label{eqn:mtpp}
\end{equation}

where \(\tau\) means integrating over time. $F^*(t)$ is the conditional \acrshort*{pdf} that no event has ever happened up to time $t$ since $t_l$. We explain how to obtain \cref{eqn:mtpp} from \cref{eqn:1} in \cref{appendix:mtpp_pdf}.

The simplest form of \acrshort*{mtpp} is the homogeneous Poisson process whose \acrshort*{cif} merely contains a positive number, i.e., \(\lambda^*(m = k_i, t) = c\). Another example is the Hawkes process \citep{hawkes_spectra_1971}, belonging to the self-exciting point process family. Its \acrshort*{cif} is $\lambda^*(m = k_i, t) = \mu_i + \sum_{j:t_j<t}{\kappa_i(t, t_j)}$ where $\kappa_i(t, t_j) > 0$ represents the excite from previous events. Because it meets the real-world intuition that the influence of occurred events always drastically drops as time passes, the Hawkes process is a widely used backbone process in various models \citep{arastuie_chip_2020, okawa_dynamic_2021, ide_cardinality-regularized_2021, huang_mutually_2022}. Recently, we have witnessed a rapid growth of neural \acrshort*{mtpp}, which models \(p^*(m, t)\) using neural networks \citep{mei_neural_2017, omi_fully_2019, zhang_self-attentive_2020, mei_transformer_2021, zhouAutomaticIntegrationSpatiotemporal2023}, due to the capability of learning complicated temporal patterns and computational efficiency \citep{Shchur2021}.

Based on \(p^*(m, t)\), the mark \(m\) and time \(t\) of the next event can be predicted. Most existing \acrshort*{mtpp} methods predict when the next event will occur first, and then predict what the mark is at the predicted time. Specifically, the expected time of the next event is \(\bar{t}=\int_{t=t_l}^{\infty} \tau p^*(\tau)d\tau\) where \(p^*(t) = \sum_{m \in M}{p^*(m, t)}\). A numerical method is typically used to calculate \(\bar{t}\) by sampling \(N\) times, denoted as \( \{ t^i\}_N \), from \(p^*(t)\) following \textit{\acrfull*{ta}} or \textit{\acrfull*{its}} \citep{rasmussen_lecture_2018} so that \(\bar{t}=\frac{1}{N}\sum_{i}{t^i}\). After that, the mark of the next event at \(\bar{t}\) is predicted: \(m_{\bar{t}} = \argmax_{m \in M}{p^*(m, \bar{t})}\). Some studies predict the mark of the next event \(m = \argmax_{m \in M}{p^*(m)}\) and then predict the time of the next event \(\bar{t}_m=\int_{t=t_l}^{\infty} \tau p^*(\tau|m)d\tau\) given the predicted mark \citep{waghmare_modeling_2022}.

\begin{figure}[ht]
\centering
\begin{minipage}{0.25\textwidth}
    \captionsetup{type=table}
    \captionof{table}{Mark prediction performance measured by macro-F1 using SAHP \citep{zhang_self-attentive_2020} for rare and frequent marks on three real-world datasets.}
    \resizebox{\linewidth}{!}{
    \begin{tabular}{llc}
        \toprule
        \multirow{4}{*}{\rotatebox[origin=c]{90}{Retweet}} & \multirow{2}{*}{\makecell{Rare\\Marks}} & \multirow{2}{*}{0.0266\tiny{$\pm$0.0135}} \\
        \\
        & \multirow{2}{*}{\makecell{Freq\\Marks}} & \multirow{2}{*}{0.6183\tiny{$\pm$0.0010}} \\
        \\
        \midrule
        \multirow{6}{*}{\rotatebox[origin=c]{90}{USearthquake}} & \multirow{3}{*}{\makecell{Rare\\Marks}} & \multirow{3}{*}{0.0037\tiny{$\pm$0.0010}} \\
        \\
        \\
        & \multirow{3}{*}{\makecell{Freq\\Marks}} & \multirow{3}{*}{0.2196\tiny{$\pm$0.0016}} \\
        \\
        \\
        \midrule
        \multirow{6}{*}{\rotatebox[origin=c]{90}{StackOverflow}} & \multirow{3}{*}{\makecell{Rare\\Marks}} & \multirow{3}{*}{0.0863\tiny{$\pm$0.0032}} \\
        \\
        \\
        & \multirow{3}{*}{\makecell{Freq\\Marks}} & \multirow{3}{*}{0.2054\tiny{$\pm$0.0011}} \\
        \\
        \\
        \bottomrule
    \end{tabular}}
    \label{tab:pred_wo_imb}
\end{minipage}
\hfill
\nextfloat
\begin{minipage}{0.7\textwidth}
    \captionsetup[subfigure]{justification=centering}
    \begin{subfigure}[b]{\textwidth}
    \begin{subfigure}{0.3\textwidth}
        \includegraphics[width=\textwidth]{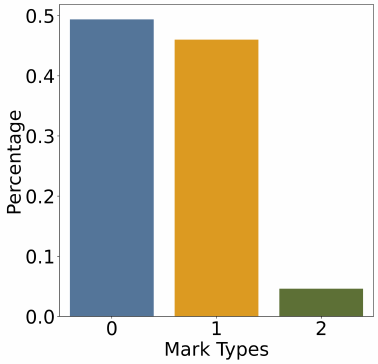}
        \label{fig:te_dataset_Retweet}
    \end{subfigure}
    \begin{subfigure}{0.3\textwidth}
        \includegraphics[width=\textwidth]{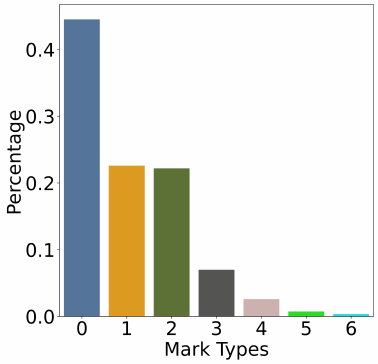}
        \label{fig:te_dataset_usearthquake}
    \end{subfigure}
    \begin{subfigure}{0.3\textwidth}
        \includegraphics[width=\textwidth]{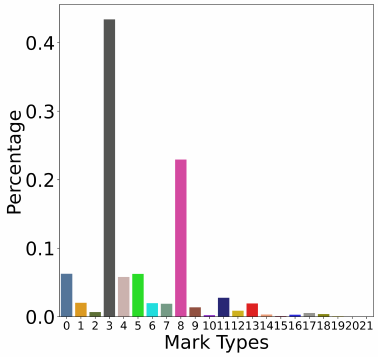}
        \label{fig:te_dataset_StackOverflow}
    \end{subfigure}
    \caption{The frequency distribution of marks in Retweet, USearthquake, and StackOverflow (from left to right).}
\end{subfigure}
\begin{subfigure}[b]{\textwidth}
    \begin{subfigure}{0.3\textwidth}
        \includegraphics[width=\textwidth]{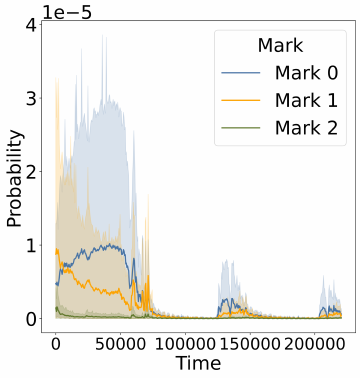}
    \end{subfigure}
    \begin{subfigure}{0.3\textwidth}
        \includegraphics[width=\textwidth]{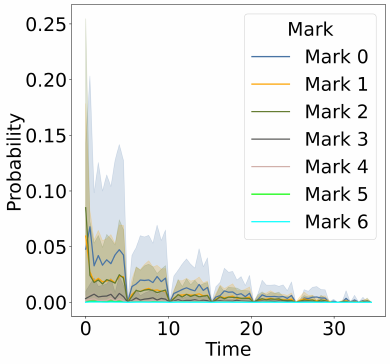}
    \end{subfigure}
    \begin{subfigure}{0.3\textwidth}
        \includegraphics[width=\textwidth]{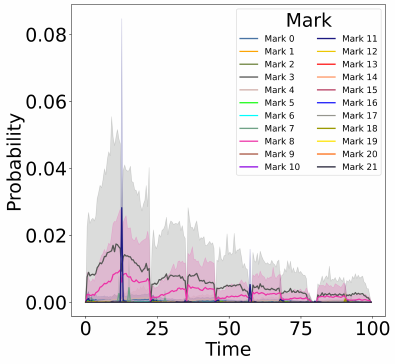}
    \end{subfigure}
    \caption{The \(p^*(m, t)\) for each mark $m$ in Retweet, USearthquake, and StackOverflow (from left to right).}
\end{subfigure}
\caption{A demonstration of the mark imbalance in various datasets and its influence on \(p^*(m, t)\).}
\label{fig:1}
\end{minipage}
\end{figure}

\subsection{Mark Imbalance} 
In real-world scenarios, the mark distribution can be significantly imbalanced, i.e., some marks are persistent and others are rare. The imbalance hurts the performance of the next event prediction, especially for rare marks, which are often more important than other marks (e.g., the occurrence of a 7-magnitude earthquake). Let us consider two marks \(k_1\) and \(k_2\) where \(k_1\) is much more frequent than \(k_2\) in the observed event sequence. Suppose the next event is mark \(k_2\). Because \(k_1\) is much more frequent than \(k_2\), it is very likely that \(p^*(k_1, t) > p^*(k_2, t)\) for most of the time \(t\), including \(\bar{t}=\int_{t=t_l}^{\infty} \tau\sum_{m\in M}p^*(m,\tau)d\tau\). If so, \(k_1\) will be predicted as the next event, but the real mark is \(k_2\).

\cref{fig:1} (a) demonstrates the mark frequency distribution in three datasets, Retweet, USearthquake, and StackOverflow (see details in \cref{sec:experiments}). \cref{fig:1} (b) shows \(p^*(m, t)\) for each mark \(m\) in these datasets. The envelope covers \(p^*(m, t)\) of all instances in the datasets, and the line is the average of \(p^*(m, t)\) across these instances. These figures show that \(p^*(k_1, t) > p^*(k_2, t)\) for most of the time if \(k_1\) is frequent and \(k_2\) is rare. \cref{tab:pred_wo_imb} shows the mark prediction performance achieved by SAHP \citep{zhang_self-attentive_2020}, a neural \acrshort*{mtpp} model, on rare and frequent marks, respectively, measured by macro-F1. We can see that the prediction performance for rare marks is significantly lower than that for frequent marks.

\section{Methodology}\label{sec:ETparadigm}
To improve the performance of \acrshort*{mtpp} for next event prediction, this study handles mark imbalance with a thresholding method, which learns thresholds to tune the mark probability normalized by the mark's prior probability to optimize mark prediction. In conjunction with the thresholding method, the proposed method predicts the mark based on $p^*(m)$ and then, given the predicted mark $m$, predicts the time based on $p^*(t|m)$.

\subsection{Next Event Mark Prediction with Thresholding}
\label{sec:thresholding}
The mark prediction of the next event depends on accurately modeling \(p^*(m)\) for each mark \(m\), the probability that the mark of the next event is \(m\), based on \(p^*(m, t)\). The expression of \(p^*(m)\) is:
\begin{equation}
    p^*(m) = \int_{t_l}^{+\infty}{p^*(m, \tau)d\tau}
\label{eqn:target1}
\end{equation}
In general, the frequent mark has a high \(p^*(m)\) and the rare mark has a low \(p^*(m)\). Inspired by the thresholding method \citep{lawrenceNeuralNetworkClassification2012, BUDA2018249}, we normalize the probability for each mark by its prior probability and learn to tune it to improve the prediction performance for rare marks. Specifically, for mark \(m\), we calculate the ratio between the probability of \(m\), \(p^*(m)\), and its prior probability, \(\overline{p}^*(m)\):
\begin{equation}
    \label{eqn:get_ratio}
    r_{m} = \frac{p^*(m)}{\overline{p}^*(m)}
\end{equation}
In this paper, \(\overline{p}^*(m)\) is the proportion of mark \(m\) in the training set. \(p^*(m)\) measures the probability that the next event mark is \(m\). If \(m\) is more frequent than \(m'\), \(p^*(m)\) is expected to be higher than \(p^*(m')\). In contrast, \(r_{m}\) evaluates whether \(p^*(m)\) is higher relative to its own proportion. A rare mark having a low \(p^*(m)\) may have a high \(r_{m}\) to signal a high chance of being the next event mark. In this way, the chance of rare marks in the next event prediction is rectified.

With \(r_{m}\) for every mark \(m\), the next event mark prediction \(m_p\) is obtained by a thresholding method:
\begin{equation}
    \label{eqn:get_mark}
    m_p = \argmax_{m}{(r_{m} - \epsilon_{m})}
\end{equation}
where \(\epsilon_m\) is the threshold of \(r_m\). 
We learn \(\epsilon_m\), which maximizes the accuracy of \(m_p\) on the training set. Specifically, for each mark \(m\), if we predict the next mark as \(m\) when \(r_m > \epsilon_m\) and not \(m\) when \(r_m \leqslant \epsilon_m\), the learned \(\epsilon_{m}\) maximizes the F1 score of the pairwise comparison, i.e., mark \(m\) vs. all other marks. The technical details of the thresholding method are in \cref{app:thresholding}.

\subsection{Next Event Time Prediction}\label{sec:netp}
After mark prediction, we predict the time. Let \(p^*(t|m)\) be the \acrshort*{pdf} of the next event time on the condition that the next event mark is \(m\). Based on \(p^*(t|m)\), we have:
\begin{equation}
\bar{t}_m = \mathds{E}_{t\sim p^*(\tau|m)}[t] = \int_{t_l}^{+\infty}{\tau p^*(\tau|m)d\tau}
\label{eqn:target}
\end{equation}
where \(\bar{t}_m\) is the expected time of the next event given the mark \(m\).

\subsection{Unifying Integral Functions}\label{sec:application}
By the definition in \cref{eqn:target1} and \cref{eqn:target}, we must solve the improper integration of \(p^*(m, \tau)\) and \(\tau p^*(\tau|m)\) for mark probability \(p^*(m)\) and time prediction \(\bar{t}\), respectively. In general, improper integration does not have analytic solutions. This means that directly calculating \(p^*(m)\) and \(\bar{t}_m\) following \cref{eqn:target1} and \cref{eqn:target} is impossible. The solution is to approximate \(p^*(m)\) and \(\bar{t}\). In particular, \(\bar{t}\) is approximated as the average of \(N\) samples \(\{t^i\}^m_N\) from $p^*(t|m)$ as \cref{eqn:pred}.
\begin{equation}
\label{eqn:pred}
\bar{t}_m = \mathds{E}_{t\sim p^*(\tau|m)}[t] \approx \frac{1}{N}{\sum_{i = 1}^{N}{t^i}}
\end{equation}
To draw \(\{t^i\}^m_N\) from $p^*(t|m)$, we use \acrfull*{its}, which takes the \acrfull*{cdf} of the distribution that one wants to sample from. In our case, let \(F^*(t|m)\) be the CDF of \(p^*(t|m)\), i.e., \(F^*(t|m)=\int_{t_l}^{t}{p^*(\tau|m)d\tau}\). \(F^*(t|m)\) refers to the probability of the next event happening in \((t_l, t]\) on the condition that its mark is \(m\). To draw a sample \(t^i\) from $p^*(t|m)$, we need to solve \cref{eqn:its}.
\begin{equation}
\label{eqn:its}
    F^*(t^i|m) = u^i
\end{equation}
where \(u^i\) is a random sample from a uniform distribution \(U(0,I)\). Since \(F^*(t|m)\) is monotonic, \cref{eqn:its} is solvable by the bisection method. For each mark \(m\), we obtain \(\{t^i\}^m_N\) by solving \cref{eqn:its} \(N\) times, which allows acquiring an arbitrary number of samples for time prediction no matter rare or frequent the mark \(m\) is. We can express \(F^*(t|m)\) as follows: 
\begin{equation}
\label{eqn:time_dist_cond_m}
    F^*(t|m) = \frac{F^*(m, t)}{p^*(m)} = \frac{1}{\int_{t_l}^{+\infty}{p^*(m, \tau)d\tau}}\int_{t_l}^{t}{p^*(m, \tau)d\tau}
\end{equation}
where \(p^*(m)=\int_{t_l}^{+\infty}{p^*(m, \tau)d\tau}\) is the probability that the mark of next event is \(m\) since \(t_l\), and \(F^*(m, t)=\int_{t_l}^{t}{p^*(m, \tau)d\tau}\) is the probability that the next event is mark \(m\) and happens in time interval \((t_l, t]\). We can further breakdown \(F^*(m, t)\) as shown in \cref{eqn:breakdown}.
\begin{equation}
\label{eqn:breakdown}
    F^*(m, t) = \int_{t_l}^{t}{p^*(m, \tau)d\tau}  
    = \int_{t_l}^{+\infty}{p^*(m, \tau)d\tau} - \int_{t}^{+\infty}{p^*(m, \tau)d\tau}
\end{equation}
For each mark $m\in \mathrm{M}$, we define $\Gamma^*(m, t)$ as the integration starting from time \(t\), any time after $t_l$ or $t_l$, to positive infinity:
\begin{equation}
    \label{eqn:gamma_def}
    \Gamma^*(m, t) = \int_{t}^{+\infty}{p^*(m, \tau)d\tau}
\end{equation}
$\Gamma^*(m, t)$ is monotonically decreasing as its derivative \(-p^*(m, t)\) is always smaller than 0. We rewrite \(p^*(m)\) in \cref{eqn:target1} and \(F^*(t|m)\) in \cref{eqn:time_dist_cond_m} using $\Gamma^*(m, t)$:
\begin{align}
    p^*(m) &= \Gamma^*(m, t_l) \\
    F^*(t|m) &= \frac{\Gamma^*(m, t_l) - \Gamma^*(m, t)}{\Gamma^*(m, t_l)}
\end{align}
This means two improper integrations in \cref{eqn:target1} and \cref{eqn:target} are now unified into one, i.e., \(\Gamma^*(m, t)\), for modeling $p^*(m)$ and time prediction.


While drawing samples from a distribution can follow \acrfull*{ta} or \acrfull*{its} \citep{rasmussen_lecture_2018}, only \acrshort*{its} is suitable for integral function unification here. The basic idea in \acrshort*{its} is to simulate using CDF of \(p^*(t|m)\). Instead, \acrfull*{ta} explicitly requires the expression of \(p^*(t|m)\), which is unknown typically. 

\mysubsection[IFNMTPP]{\acrfull*{model}}
\label{sec:model}
With \(\Gamma^*(m, t)\), we can model \(p^*(m)\) and prediction time \(\bar{t}_m\). However, \(\Gamma^*(m, t)\) is an improper integral with an infinite integration interval. Numerical methods are computationally expensive and can only be used to estimate integrals on a finite interval. To avoid numerical methods, we introduce \acrfull*{model} to approximate \(\Gamma^*(m, t)\). For each mark $m$, \acrshort*{model} models the relationship between $p^*(m,t)$ and its integral \(\Gamma^*(m,t)\).

\begin{figure*}[ht]
    \centering
    \includegraphics[width=\linewidth]{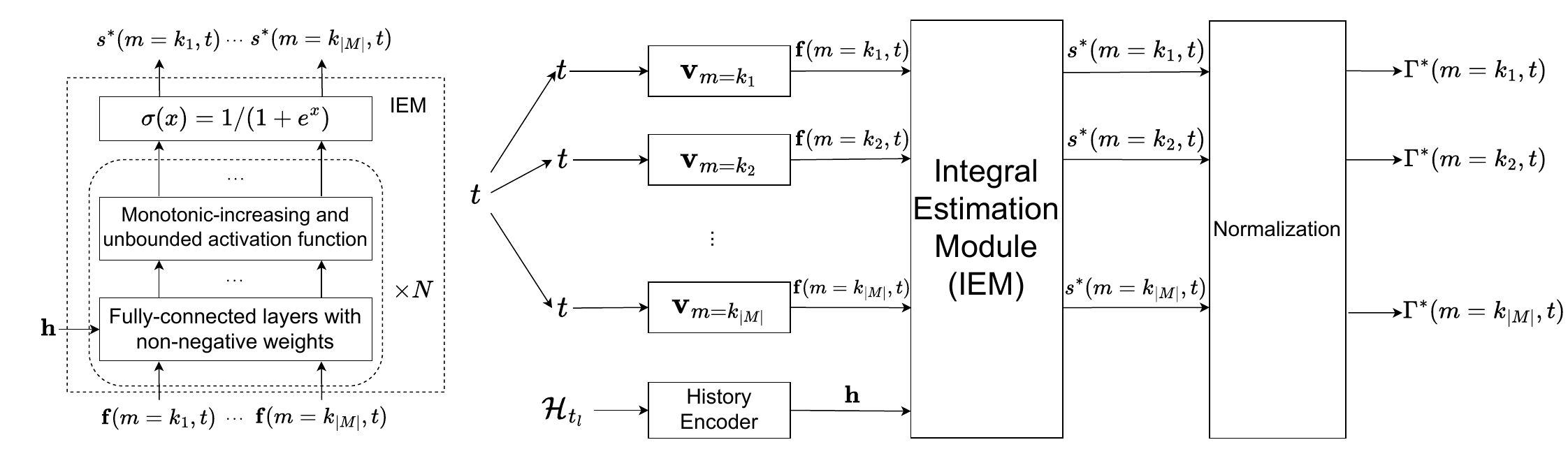}
    \caption{Architecture of \acrshort*{model} where the history encoder is an LSTM.}
    \label{fig:IFIB}
\end{figure*}

\cref{fig:IFIB} sketches the architecture of \acrshort*{model}. For each mark $m\in \mathrm{M}$, we assign a vector $\mathbf{v}_m$ to prepare $\mathbf{f}(m, t)=\mathbf{v}_{m}(t - t_l) + \mathbf{b}_m$ as input of the \acrfull*{iem}. All parameters in \(\mathbf{v}_{m}\) are non-negative.
\acrshort*{iem} contains multiple fully-connected layers with non-negative weights, and monotonic-increasing and unbounded activation functions. Then, it ends with a monotonically decreasing function $\sigma(x) = 1/(1 + e^x)$. So, IFNMTPP is intrinsically monotonically decreasing w.r.t. \(t\). The outputs of \acrshort*{iem} are scores $s^*(m = k_1, t), s^*(m = k_2, t), \cdots, s^*(m = k_{|\mathrm{M}|}, t)$. The value of $\sum_{m\in \mathrm{M}}s^*(m,t)$ is not guaranteed to be 1. To produce the qualified probability distribution, they need to be normalized. This is achieved by the Normalization module in \cref{fig:IFIB} that divides $s^*(m, t)$ by the partition function $Z(\mathcal{H}_{t_l}) = \sum_{m \in \mathrm{M}}{s^*(m, t_l)}$ for each $m\in M$. Finally, \acrshort*{model} outputs $\Gamma^*(m,t)$ for each mark $m$ at the given time $t$:

\begin{equation}
\label{eqn:IFIB_1}
    \Gamma^*(m, t) = \frac{s^*(m, t)}{Z(\mathcal{H}_{t_l})}
\end{equation}

With $\Gamma^*(m, t)$ and $\Gamma^*(m, t_l)$, we have \(F^*(t|m)\) by \cref{eqn:time_dist_cond_m} and \cref{eqn:breakdown}. Next, we calculate \(\bar{t}_m\) by drawing \(\{t^i\}^m_N\) from \(F^*(t|m)\) following \cref{eqn:its}. With the definition of \acrshort*{model}, we have the following proposition, with the proof in \cref{appendix:ifnmtpp_correct}, to guarantee that the model output is \(\Gamma^*(m, t)\):
\begin{proposition}
    The output of \acrshort*{model} is \(\Gamma^*(m, t)\) when its gradient is \(-p^*(m, t)\).
    \label{prop:ifnmtpp_correct}
\end{proposition}

We train \acrshort*{model} using the Negative Log-Likelihood (NLL) on event sequence \(\mathcal{S}\) observed in a time interval \([t_0, T)\), where the time of the first event is $t_1 \geq t_0$, and the time of the last event is $t_s \leq T$.
\begin{equation}
\begin{aligned}
L =& -\log p(\mathcal{S}) = -\sum_{(m_i, t_i) \in \mathcal{S}}{\log\lambda^*(m_i, t_i)} + \int_{t_0}^{T}{\sum_{n \in M}{\lambda^*(n, \tau)d\tau}} \\
=& - \sum_{(m_i, t_i) \in \mathcal{S}}{(\log\lambda^*(m_i, t_i)} - \int_{t_{i-1}}^{t_i}{\sum_{n \in M}{\lambda^*(n, \tau)d\tau})} + \int_{t_s}^{T}{\sum_{n \in M}{\lambda^*(n, \tau)d\tau})} \\
=& - \sum_{(m_i, t_i) \in \mathcal{S}}{\log p^*(m_i, t_i)} - \log(1 - \sum_{n \in M}{F^*(n, T)}) \\
=& - \sum_{(m_i, t_i) \in \mathcal{S}}{\log p^*(m_i, t_i)} - \log(\sum_{n \in M}{\Gamma^*(n, T)}) \\
\end{aligned}
\label{eqn:mtpp_loss}
\end{equation}
where $p^*(m_i,t_i)$ is the probability of the $i$th event conditioned on historical events.
IFNMTPP increases $p^*(m_i,t_i)$ where $(m_i,t_i)\in \mathcal{S}$ are real events in event sequences. The term \(\log (\sum_{m \in M}{\Gamma^*(m, T)})\) is the survival term, 
which models no events after the last event $t_s$ in each event sequence until time $T$.
In \acrshort*{model}, the expression of $p^*(m_i,t_i)$ is:
\begin{equation}
\label{eqn:IFIB_2}
    p^*(m_i,t_i)
    = -\frac{\partial \Gamma^*(m_i,t_i)}{\partial t_i} = -\frac{1}{Z(\mathcal{H}_{t_l})}\frac{\partial s^*(m_i,t_i)}{\partial \mathbf{f}(m_i,t_i)}\frac{\partial \mathbf{f}(m_i,t_i)}{\partial t_i}
\end{equation}
\citet{mei_noise-contrastive_2020} prove that an MTPP model converges to the true distribution when trained with the NLL loss defined in \cref{eqn:mtpp_loss}. Combined with \cref{prop:ifnmtpp_correct}, \acrshort*{model} consistently estimates the true value of \(\Gamma^*(m, t)\).

\section{Experiments}\label{sec:experiments}
We run every experiment 5 times with different random seeds and report the mean and standard deviation (1-sigma) of all results. The complete experiment settings are described in \cref{app:hyperparameters}.

\textbf{Datasets}\footnote{Retweet, StackOverflow, Taobao, and USearthquake are released under Apache-2.0 license\citep{xue_easytpp_2023}.} Four real-world datasets include Retweet \citep{zhao_seismic_2015}, StackOverflow(SO) \citep{Leskovec2014SNAPD}, Taobao User Behavior Data(Taobao) \citep{taobao}, and earthquake events over the Conterminous US(USearthquake) \citep{xue_easytpp_2023}. We split all marks of each dataset into two subsets, one containing frequent marks, denoted as \(\mathrm{M}_f\), the other containing rare marks, denoted as \(\mathrm{M}_r\). \(\mathrm{M}_r \cap \mathrm{M}_f = \varnothing\) and \(\mathrm{M}_r \cup \mathrm{M}_f = \mathrm{M}\). The rare marks and frequent marks for each dataset are described in \cref{apx:datapre}.

\textbf{Baseline Models}\footnote{Our codes will be released under MIT license.} Our method, denoted as \textit{ours}, uses \acrshort*{model} for predicting the mark of the next event, optimized with thresholding, then uses \acrshort*{model} to predict the time of the next event given the predicted mark. The first group of baselines includes: (i) \textit{ours-w/o-thresholding} to evaluate the effectiveness of the thresholding method. (ii) \textit{time-mark-with-thresholding} to evaluate the necessity to predict marks first for handling mark imbalance with thresholding. (iii) \textit{time-mark-w/o-thresholding} same as time-mark-with-thresholding but mark prediction is not optimized with thresholding. The second group of baselines evaluates thresholding against resampling, another classic technique to address data imbalance, including \textit{undersampling} and \textit{oversampling}. The third group of baselines includes existing \acrshort*{mtpp} methods. Since \acrshort*{mtpp} modeling has been well studied in the past decades, the state-of-the-art methods demonstrate comparable performance. Among them, this study selects the most popular ones as baselines, including FullyNN \citep{omi_fully_2019}, THP \citep{zuo_transformer_2020}, SAHP \citep{zhang_self-attentive_2020}, AttNHP \citep{mei_transformer_2021}, and Marked-LNM \citep{waghmare_modeling_2022}. The details of these baselines are available in \cref{app:baseline}.

\textbf{Evaluation Metrics}
We use macro-F1 and micro-F1, described in \cref{app:mi-ma-f1}, to evaluate mark predictions and use \acrfull*{mmae}, described in \cref{app:mae}, to evaluate time predictions on real-world datasets. The evaluation metrics for mark and time are independent of each other. For each dataset, as discussed above, three sets \(\mathrm{M}\), \(\mathrm{M}_f\), and \(\mathrm{M}_r\) are drawn from the original test set. Moreover, we evaluate the fidelity of \acrshort*{model} using five synthetic datasets. Experiment results on synthetic datasets (reported in \cref{app:synthetic}) demonstrate the high fidelity of \acrshort*{model} compared with other \acrshort*{mtpp} models. 

\begin{table*}[!t]
    \caption{Mark prediction performance measured by macro-F1/micro-F1 to evaluate thresholding and the prediction order of mark and time. The bold are the best values.}
   \begin{small}
   \setlength{\tabcolsep}{2pt}
   \resizebox{\linewidth}{!}{
   \begin{tabular}{llcccc}
       \toprule
                                & & Retweet & SO & Taobao & USearthquake \\
       \midrule
       \multirow{3}{*}{\makecell{ours}}   & \(\mathrm{M}\) & \textbf{0.4750\tiny{$\pm$0.0033}}/\textbf{\meanstd{0.4394}{0.0093}} & \textbf{0.1776\tiny{$\pm$0.0030}}/\textbf{\meanstd{0.6376}{0.0026}} & \textbf{0.4190\tiny{$\pm$0.0104}}/\textbf{\meanstd{0.7499}{0.0151}} & \textbf{0.1382\tiny{$\pm$0.0071}}/\textbf{\meanstd{0.3189}{0.0125}} \\
       & \(\mathrm{M}_r\) & \textbf{0.2010\tiny{$\pm$0.0082}}/\textbf{\meanstd{0.2010}{0.0082}} & \textbf{0.1476\tiny{$\pm$0.0041}}/\textbf{\meanstd{0.4530}{0.0026}} & \textbf{0.3987\tiny{$\pm$0.0108}}/\textbf{\meanstd{0.7558}{0.0185}} & \textbf{0.0339\tiny{$\pm$0.0051}}/\textbf{\meanstd{0.1111}{0.0098}} \\
       & \(\mathrm{M}_f\) & \textbf{0.6120\tiny{$\pm$0.0013}}/\meanstd{0.9612}{0.0021} & {0.2795\tiny{$\pm$0.0014}}/\meanstd{0.8974}{0.0042} & \textbf{0.7441\tiny{$\pm$0.0060}}/\textbf{\meanstd{0.7441}{0.0060}} & \textbf{0.2773\tiny{$\pm$0.0215}}/\meanstd{0.9181}{0.0102} \\
       
       \midrule
       \multirow{3}{*}{\makecell[l]{time-mark-\\with-thresholding}}   & \(\mathrm{M}\) & {0.4741\tiny{$\pm$0.0028}}/\meanstd{0.4380}{0.0016} & {0.1431\tiny{$\pm$0.0075}}/\meanstd{0.5834}{0.0028} & {0.3289\tiny{$\pm$0.0193}}/\meanstd{0.7059}{0.0384} & {0.1214\tiny{$\pm$0.0126}}/\meanstd{0.3091}{0.0083} \\
       & \(\mathrm{M}_r\) & {0.2000\tiny{$\pm$0.0067}}/\meanstd{0.2000}{0.0011} & {0.1023\tiny{$\pm$0.0094}}/\meanstd{0.3829}{0.0046} & {0.3054\tiny{$\pm$0.0201}}/\meanstd{0.7072}{0.0690} & {0.0298\tiny{$\pm$0.0093}}/\meanstd{0.1049}{0.0059} \\
       & \(\mathrm{M}_f\) & {0.6093\tiny{$\pm$0.0008}}/\meanstd{0.9596}{0.0019} & \textbf{0.2815\tiny{$\pm$0.0022}}/\meanstd{0.8888}{0.0041} & {0.7062\tiny{$\pm$0.0204}}/\meanstd{0.7062}{0.0204} & {0.2436\tiny{$\pm$0.0236}}/\meanstd{0.9116}{0.0064} \\

       \midrule
       \multirow{3}{*}{\makecell[l]{ours-\\w/o-thresholding}} & \(\mathrm{M}\) & {0.4269\tiny{$\pm$0.0010}}/\meanstd{0.1800}{0.0093} & {0.1287\tiny{$\pm$0.0031}}/\meanstd{0.5877}{0.0023} & {0.3968\tiny{$\pm$0.0138}}/\meanstd{0.7183}{0.0220} & {0.1153\tiny{$\pm$0.0061}}/\meanstd{0.1172}{0.0048} \\
       & \(\mathrm{M}_r\) & {0.0333\tiny{$\pm$0.0082}}/\meanstd{0.0333}{0.0034} & {0.1065\tiny{$\pm$0.0047}}/\meanstd{0.3763}{0.0030} & {0.3759\tiny{$\pm$0.0150}}/\meanstd{0.7059}{0.0332} & {0.0121\tiny{$\pm$0.0035}}/\meanstd{0.0146}{0.0012} \\
       & \(\mathrm{M}_f\) & {0.6238\tiny{$\pm$0.0001}}/\textbf{\meanstd{0.9770}{0.0000}} & {0.2043\tiny{$\pm$0.0022}}/\textbf{\meanstd{0.9180}{0.0004}} & {0.7311\tiny{$\pm$0.0108}}/\meanstd{0.7311}{0.0108} & {0.2528\tiny{$\pm$0.0118}}/\textbf{\meanstd{0.9451}{0.0005}} \\

       \midrule
       \multirow{3}{*}{\makecell[l]{time-mark-\\w/o-thresholding}}   & \(\mathrm{M}\) & {0.4252\tiny{$\pm$0.0033}}/\meanstd{0.1815}{0.0077} & {0.0906\tiny{$\pm$0.0055}}/\meanstd{0.4428}{0.0080} & {0.3135\tiny{$\pm$0.0136}}/\meanstd{0.6384}{0.0316} & {0.1066\tiny{$\pm$0.0040}}/\meanstd{0.1111}{0.0336} \\
       & \(\mathrm{M}_r\) & {0.0338\tiny{$\pm$0.0029}}/\meanstd{0.0338}{0.0029} & {0.0567\tiny{$\pm$0.0071}}/\meanstd{0.2142}{0.0076} & {0.2897\tiny{$\pm$0.0137}}/\meanstd{0.5884}{0.0467} & {0.0033\tiny{$\pm$0.0024}}/\meanstd{0.0143}{0.0092} \\
       & \(\mathrm{M}_f\) & {0.6208\tiny{$\pm$0.0008}}/\textbf{\meanstd{0.9770}{0.0001}} & {0.2059\tiny{$\pm$0.0004}}/\meanstd{0.9156}{0.0013} & {0.6933\tiny{$\pm$0.0129}}/\meanstd{0.6933}{0.0129} & {0.2444\tiny{$\pm$0.0085}}/\meanstd{0.9446}{0.0002} \\
   \bottomrule
   \end{tabular}}
   \end{small}
   \label{tab:thresholding}
\end{table*}

\subsection{Experiment results}

\textbf{Impact of Thresholding on Mark Prediction}\quad
For the mark prediction, we evaluate (i) the effectiveness of the proposed thresholding method by comparing \textit{ours} with \textit{ours-w/o-thresholding}, (ii) the necessity of the strategy to predict the mark first by comparing \textit{ours} with \textit{time-mark-with-thresholding}. The metric is macro-F1 and micro-F1, where the higher values indicate more accurate mark predictions. The experimental results are reported in \cref{tab:thresholding}. Compared with \textit{ours-w/o-thresholding}, \textit{ours} performs much better on rare and all mark prediction. \textit{ours} also shows a comparable performance on frequent mark prediction. It implies that the recall of frequent mark prediction is improved using thresholding. 

In \cref{tab:thresholding}, the performance of \textit{ours} is better than \textit{time-mark-with-thresholding} on rare mark prediction on all datasets. To check whether the suboptimal performance of \textit{time-mark-with-thresholding} is due to applying thresholding or not, we also compare \textit{time-mark-with-thresholding} with \textit{time-mark-w/o-thresholding}. The experimental results show that the performance of \textit{time-mark-w/o-thresholding} is lower than that of \textit{time-mark-with-thresholding}. This indicates that predicting the mark first is more suitable for handling mark imbalance with thresholding.

\begin{wraptable}{r}{70mm}
    \vskip -0.1in
    \captionsetup{type=table, hypcap=false}
    \caption{Time prediction performance to evaluate the order of time and mark predictions. The bold are the best values.}
    \resizebox{\linewidth}{!}{
    \setlength{\tabcolsep}{1pt}
    \label{tab:p_t_m_p_t}
    \begin{tabular}{llcccc}
        \toprule
        &  &  Retweet  &  SO  &  Taobao  &  USearthquake  \\ 
        \midrule
        \multirow{3}{*}{\makecell{{ours}}} & \(\mathrm{M}\) & {2515.1\tiny{$\pm$6.5029}} & \textbf{0.5212\tiny{$\pm$0.0142}} & {0.3324\tiny{$\pm$0.0579}} & \textbf{0.6856\tiny{$\pm$0.0063}} \\
        & \(\mathrm{M}_{r}\) & {3291.2\tiny{\(\pm\)29.097}} &\textbf{0.4986\tiny{\(\pm\)0.0175}} & {0.3385\tiny{\(\pm\)0.0627}} &\textbf{0.6966\tiny{\(\pm\)0.0081}} \\
        & \(\mathrm{M}_{f}\) & \textbf{2198.7\tiny{\(\pm\)2.4798}} & \textbf{0.6063\tiny{\(\pm\)0.0001}} & {0.2529\tiny{\(\pm\)0.0055}} & \textbf{0.6713\tiny{\(\pm\)0.0048}} \\
        \midrule
        \multirow{3}{*}{\makecell[l]{time-mark-\\with-\\thresholding}} & \(\mathrm{M}\) & \textbf{2504.5\tiny{\(\pm\)4.4738}} & {0.6417\tiny{\(\pm\)0.0127}} & \textbf{0.2420\tiny{\(\pm\)0.0227}} & {0.8516\tiny{\(\pm\)0.2378}} \\
        & \(\mathrm{M}_{r}\) & \textbf{3223.2\tiny{\(\pm\)10.354}} & {0.6515\tiny{\(\pm\)0.0167}} & \textbf{0.2411\tiny{\(\pm\)0.0215}} & {0.7008\tiny{\(\pm\)0.0075}} \\
        & \(\mathrm{M}_{f}\) & {2207.7\tiny{\(\pm\)3.1220}} & {0.6095\tiny{\(\pm\)0.0008}} & \textbf{0.2573\tiny{\(\pm\)0.0429}} & {1.2295\tiny{\(\pm\)0.7947}} \\
        \bottomrule
    \end{tabular}}
        \caption{Time prediction performance to evaluate thresholding vs. resampling. The bold are the best.}
        \resizebox{\linewidth}{!}{
        \setlength{\tabcolsep}{1pt}
        \begin{tabular}{llcccc}
        \toprule
        &  &  Retweet  &  SO  &  Taobao  &  USearthquake  \\ 
        \midrule
        \multirow{3}{*}{\makecell{{ours}}} & \(\mathrm{M}\) & {2515.1\tiny{$\pm$6.5029}} & \textbf{0.5212\tiny{$\pm$0.0142}} & \textbf{0.3324\tiny{$\pm$0.0579}} & \textbf{0.6856\tiny{$\pm$0.0063}} \\
        & \(\mathrm{M}_{r}\) & {3291.2\tiny{\(\pm\)29.097}} &\textbf{0.4986\tiny{\(\pm\)0.0175}} &\textbf{0.3385\tiny{\(\pm\)0.0627}} &\textbf{0.6966\tiny{\(\pm\)0.0081}} \\
        & \(\mathrm{M}_{f}\) & \textbf{2198.7\tiny{\(\pm\)2.4798}} & \textbf{0.6063\tiny{\(\pm\)0.0001}} & \textbf{0.2529\tiny{\(\pm\)0.0055}} & \textbf{0.6713\tiny{\(\pm\)0.0048}} \\
        
        \midrule
        \multirow{3}{*}{\makecell[l]{Over-\\sampling}} & \(\mathrm{M}\) & \textbf{2514.6\tiny{$\pm$7.3797}} & {6.7145\tiny{$\pm$7.8985}} & {3.2026\tiny{$\pm$0.0338}} & {10.598\tiny{$\pm$18.104}} \\
        & \(\mathrm{M}_{r}\) & 
        \textbf{3197.0\tiny{\(\pm\)6.0093}} & {3.2344\tiny{\(\pm\)3.4202}} & {3.1683\tiny{\(\pm\)0.0375}} & {10.098\tiny{\(\pm\)18.678}} \\
        & \(\mathrm{M}_{f}\) & {2230.1\tiny{\(\pm\)7.7471}} & {96.441\tiny{\(\pm\)122.71}} & {3.8056\tiny{\(\pm\)0.0389}} & {16.139\tiny{\(\pm\)19.098}} \\
        
        \midrule
        \multirow{3}{*}{\makecell[l]{Under-\\sampling}} & \(\mathrm{M}\) & {2526.9\tiny{$\pm$10.085}} & {3.5537\tiny{$\pm$3.9587}} & {3.2130\tiny{$\pm$0.0313}} & {17.086\tiny{$\pm$21.001}} \\
        & \(\mathrm{M}_{r}\) & {3216.9\tiny{\(\pm\)22.171}} &{1.8360\tiny{\(\pm\)1.4894}} &{3.1786\tiny{\(\pm\)0.0339}} &{16.326\tiny{\(\pm\)22.010}}  \\
        & \(\mathrm{M}_{f}\) & {2239.5\tiny{\(\pm\)5.7513}} & {54.502\tiny{\(\pm\)76.132}} & {3.8188\tiny{\(\pm\)0.0367}} & {26.183\tiny{\(\pm\)18.856}} \\
        \bottomrule
        \end{tabular}
        }
    \vskip -0.2in
\end{wraptable}

\textbf{Time Prediction Performance}\quad
For time prediction performance evaluation, we compare the time predicted using \textit{ours}, i.e., based on \(p^*(t|m)\), and that using \textit{time-mark-with-thresholding}, i.e., based on \(p^*(t)\). The metric is \acrshort*{mmae}. Lower \acrshort*{mmae} means better. The experimental results are reported in \cref{tab:p_t_m_p_t}. We observe that predicting time based on \(p^*(t|m)\) slightly outperforms that based on \(p^*(t)\). The results indicate that the strategy to predict the mark first and then time also benefits the time prediction. The reason could be that \(\bar{t}_m\) obtained by drawing samples from \(p^*(t|m)\) is more specific to the mark and thus tends to be more accurate compared to \(\bar{t}\) obtained based on \(p^*(t)\) for all marks.

\textbf{Thresholding vs. Resampling}\quad
We compare the prediction performance of \textit{ours} against resampling baselines \textit{oversampling} and \textit{undersampling}. As discussed in \cref{sec:introduction}, \textit{resampling the training set} and \textit{cost-sensitive approaches} are two commonly used methods for handling data imbalance besides \textit{thresholding}. According to \citet{lopezAnalysisPreprocessingVs2012}, \textit{resampling the training set} and \textit{cost-sensitive approaches} are statistically equivalent. So, we focus on \textit{resampling the training set} only. The experiment results are reported in \cref{tab:thresholding_vs_resampling}. We observe that \textit{ours} consistently outperforms \textit{oversampling} and \textit{undersampling}. It is easy to see that the resampling ratio impacts the performance, but it is hard to figure out the correct ratio for different marks on different datasets.

\begin{table*}[!ht]
   \caption{Mark prediction performance to evaluate thresholding and resampling, measured by macro-F1/micro-F1, The bold are the best values.}
   \centering
   \begin{small}
   \setlength{\tabcolsep}{2pt}
   \resizebox{\linewidth}{!}{
   \begin{tabular}{llcccc}
       \toprule
                                & & Retweet & SO & Taobao & USearthquake \\
       \midrule
       \multirow{3}{*}{\makecell{{ours}}}   & \(\mathrm{M}\) & \textbf{0.4750\tiny{$\pm$0.0033}}/\textbf{\meanstd{0.4394}{0.0093}} & \textbf{0.1776\tiny{$\pm$0.0030}}/\textbf{\meanstd{0.6376}{0.0026}} & \textbf{0.4190\tiny{$\pm$0.0104}}/\textbf{\meanstd{0.7499}{0.0151}} & \textbf{0.1382\tiny{$\pm$0.0071}}/\textbf{\meanstd{0.3189}{0.0125}} \\
       & \(\mathrm{M}_r\) & \textbf{0.2010\tiny{$\pm$0.0082}}/\textbf{\meanstd{0.2010}{0.0082}} & \textbf{0.1476\tiny{$\pm$0.0041}}/\textbf{\meanstd{0.4530}{0.0026}} & \textbf{0.3987\tiny{$\pm$0.0108}}/\textbf{\meanstd{0.7558}{0.0185}} & {0.0339\tiny{$\pm$0.0051}}/{\meanstd{0.1111}{0.0098}} \\
       & \(\mathrm{M}_f\) & \textbf{0.6120\tiny{$\pm$0.0013}}/\textbf{\meanstd{0.9612}{0.0021}} & \textbf{0.2795\tiny{$\pm$0.0014}}/\textbf{\meanstd{0.8974}{0.0042}} & \textbf{0.7441\tiny{$\pm$0.0060}}/\textbf{\meanstd{0.7441}{0.0060}} & \textbf{0.2773\tiny{$\pm$0.0215}}/\textbf{\meanstd{0.9181}{0.0102}} \\
       
       \midrule
       \multirow{3}{*}{\makecell{{Oversampling}}} & \(\mathrm{M}\) & {0.2368\tiny{$\pm$0.0197}}/\meanstd{0.3484}{0.0041} & {0.0635\tiny{$\pm$0.0184}}/\meanstd{0.4574}{0.0496} & {0.3538\tiny{$\pm$0.0063}}/\meanstd{0.7269}{0.0346} & {0.0647\tiny{$\pm$0.0165}}/\meanstd{0.2141}{0.0710} \\
       & \(\mathrm{M}_r\) & {0.1452\tiny{\(\pm\)0.0016}}/\meanstd{0.1452}{0.0016} & {0.0447\tiny{\(\pm\)0.0233}}/\meanstd{0.3330}{0.0340} &{0.3341\tiny{\(\pm\)0.0090}}/\meanstd{0.8059}{0.0001} & \textbf{0.0392\tiny{\(\pm\)0.0071}}/\textbf{\meanstd{0.1818}{0.0022}} \\
       & \(\mathrm{M}_f\) & {0.2859\tiny{\(\pm\)0.0176}}/\meanstd{0.2859}{0.0249} & {0.1272\tiny{\(\pm\)0.0001}}/\meanstd{0.6284}{0.0720} & {0.6570\tiny{\(\pm\)0.0623}}/\meanstd{0.6570}{0.0623} & {0.0988\tiny{\(\pm\)0.0477}}/\meanstd{0.2819}{0.1707} \\
       
       \midrule
       \multirow{3}{*}{\makecell{{Undersampling}}} & \(\mathrm{M}\)  & {0.2284\tiny{$\pm$0.0126}}/\meanstd{0.3230}{0.0391} & {0.0709\tiny{$\pm$0.0226}}/\meanstd{0.4171}{0.0357} & {0.3513\tiny{$\pm$0.0069}}/\meanstd{0.7239}{0.0102} & {0.0576\tiny{$\pm$0.0121}}/\meanstd{0.3067}{0.0292} \\
       & \(\mathrm{M}_r\) & {0.1422\tiny{\(\pm\)0.0170}}/\meanstd{0.1422}{0.0170} &{0.0544\tiny{\(\pm\)0.0263}}/\meanstd{0.3017}{0.0105} &{0.3328\tiny{\(\pm\)0.0090}}/\meanstd{0.8143}{0.0048} &{0.0382\tiny{\(\pm\)0.0084}}/\meanstd{0.1742}{0.0077} \\
       & \(\mathrm{M}_f\) & {0.2714\tiny{\(\pm\)0.0143}}/\meanstd{0.7338}{0.0898} & {0.1271\tiny{\(\pm\)0.0174}}/\meanstd{0.5850}{0.1189} & {0.6435\tiny{\(\pm\)0.0144}}/\meanstd{0.6435}{0.0144} & {0.0836\tiny{\(\pm\)0.0353}}/\meanstd{0.5507}{0.1273} \\
   \bottomrule
   \end{tabular}
}
   \end{small}
   \label{tab:thresholding_vs_resampling}
\vskip -0.1in
\end{table*}

\begin{wraptable}{R}{70mm}
    \centering
    \captionsetup{type=table, hypcap=false}
    \caption{Time prediction performance to evaluate \textit{ours} vs. existing \acrshort*{mtpp} models. The bold are the best values.}
    \resizebox{\linewidth}{!}{
    \setlength{\tabcolsep}{1pt}
    \begin{tabular}{llccccc}
    \toprule
    &  &  Retweet  &  SO  &  Taobao  &  USearthquake  \\ 
    \midrule
    \multirow{3}{*}{\makecell{{ours}}} & \(\mathrm{M}\) & \textbf{2515.1\tiny{$\pm$6.5029}} & \textbf{0.5212\tiny{$\pm$0.0142}} & {0.3324\tiny{$\pm$0.0579}} & \textbf{0.6856\tiny{$\pm$0.0063}} \\
    & \(\mathrm{M}_{r}\) & \textbf{3291.2\tiny{\(\pm\)29.097}} &\textbf{0.4986\tiny{\(\pm\)0.0175}} &{0.3385\tiny{\(\pm\)0.0627}} &\textbf{0.6966\tiny{\(\pm\)0.0081}} \\
    & \(\mathrm{M}_{f}\) & \textbf{2198.7\tiny{\(\pm\)2.4798}} & \textbf{0.6063\tiny{\(\pm\)0.0001}} & {0.2529\tiny{\(\pm\)0.0055}} & \textbf{0.6713\tiny{\(\pm\)0.0048}} \\
    \midrule
    \multirow{3}{*}{FullyNN} & \(\mathrm{M}\) & 5126.0\tiny{$\pm$854.88} & 0.7047\tiny{$\pm$0.0203} & 6.5079\tiny{$\pm$2.0854} & 1.2684\tiny{$\pm$0.3715} \\
    & \(\mathrm{M}_{r}\) & 7525.4\tiny{\(\pm\)1037.4} & 0.7231\tiny{\(\pm\)0.0269} & 6.6713\tiny{\(\pm\)2.1557} & 1.2709\tiny{\(\pm\)0.3356} \\
    & \(\mathrm{M}_{f}\) & 4232.0\tiny{\(\pm\)769.50} & 0.6461\tiny{\(\pm\)0.0029} & 4.4131\tiny{\(\pm\)1.3632} & 1.2671\tiny{\(\pm\)0.4082} \\
    \midrule
    \multirow{3}{*}{SAHP} & \(\mathrm{M}\) & 3320.0\tiny{$\pm$242.70} & 0.8010\tiny{$\pm$0.0593} & 23.409\tiny{$\pm$14.564} & {0.7608\tiny{$\pm$0.0588}} \\
    & \(\mathrm{M}_{r}\) & 4260.9\tiny{\(\pm\)618.21} & 0.7882\tiny{\(\pm\)0.0734} & 27.638\tiny{\(\pm\)17.438} & {0.7777\tiny{\(\pm\)0.0650}} \\
    & \(\mathrm{M}_{f}\) & 2936.3\tiny{\(\pm\)118.11} & 0.8493\tiny{\(\pm\)0.0208} & 1.7466\tiny{\(\pm\)0.8728} & {0.7388\tiny{\(\pm\)0.0512}} \\
    \midrule
    \multirow{3}{*}{THP} & \(\mathrm{M}\) & 3601.1\tiny{$\pm$231.52} & {0.6433\tiny{$\pm$0.0059}} & 3.0100\tiny{$\pm$0.2806} & 0.7322\tiny{$\pm$0.0078} \\
    & \(\mathrm{M}_{r}\) & 4250.6\tiny{\(\pm\)211.71} & {0.6586\tiny{\(\pm\)0.0080}} & 3.0036\tiny{\(\pm\)0.2893} & 0.7409\tiny{\(\pm\)0.0057} \\
    & \(\mathrm{M}_{f}\) & 3315.2\tiny{\(\pm\)241.35} & {0.6097\tiny{\(\pm\)0.0010}} & 3.3648\tiny{\(\pm\)1.0252} & 0.7207\tiny{\(\pm\)0.0114} \\
    \midrule
    \multirow{3}{*}{\makecell[l]{AttNHP}} & \(\mathrm{M}\) & {3551.1\tiny{$\pm$12.611}} & 7.9305\tiny{$\pm$5.9188} & 5.4038\tiny{$\pm$1.3280} & 6.4583\tiny{$\pm$2.2939} \\
    & \(\mathrm{M}_{r}\) & {4406.5\tiny{\(\pm\)17.518}} & 6.8682\tiny{\(\pm\)4.8605} & 5.2849\tiny{\(\pm\)1.2944} & 6.6158\tiny{\(\pm\)2.3176} \\
    & \(\mathrm{M}_{f}\) & {3187.8\tiny{\(\pm\)10.646}} & 13.197\tiny{\(\pm\)11.171} & {7.7158\tiny{\(\pm\)1.9980}} & 6.2544\tiny{\(\pm\)2.2619} \\
    \midrule
    \multirow{3}{*}{\makecell[l]{Marked-\\LNM}} & \(\mathrm{M}\) & {2559.8\tiny{$\pm$5.9380}} & 0.9067\tiny{$\pm$0.3687} & \textbf{0.2058\tiny{$\pm$0.0079}} & 0.7646\tiny{$\pm$0.0026} \\
    & \(\mathrm{M}_{r}\) & {3314.3\tiny{\(\pm\)1.2460}} & 1.0520\tiny{\(\pm\)0.5330} & \textbf{0.2043\tiny{\(\pm\)0.0091}} & 0.7773\tiny{\(\pm\)0.0057} \\
    & \(\mathrm{M}_{f}\) & {2249.7\tiny{\(\pm\)7.4050}} & 0.6084\tiny{\(\pm\)0.0007} & \textbf{0.2318\tiny{\(\pm\)0.0128}} & 0.7480\tiny{\(\pm\)0.0013} \\
    \bottomrule
    \end{tabular}
    \label{tab:IFIB_real_world_mae_e}
    }
\end{wraptable}
\textbf{Performance Comparison with Existing \acrshort*{mtpp} models}\quad \label{exp:part_1}
\cref{tab:IFIB_real_world_f1} and \cref{tab:IFIB_real_world_mae_e} report time prediction performance and mark prediction performance, respectively, of \textit{ours} and existing \acrshort*{mtpp} models, including FullyNN, THP, SAHP, AttNHP, and Marked-LNM. Compared with existing \acrshort*{mtpp} models, \textit{ours} demonstrates superior performance in both time prediction and mark prediction. For mark prediction, \textit{ours} is the first \acrshort*{mtpp} model which addresses mark imbalance. For time prediction, the time for each mark \(m\) is predicted by drawing samples from \(p^*(t|m)\) based on \(\Gamma^*(m, t)\). The accurate approximation of \(\Gamma^*(m, t)\) leads to accurate time prediction. In particular, \textit{ours} also outperforms Marked-LNM in time prediction. This demonstrates that modeling \(\Gamma^*(m, t)\) by neural networks is better than directly modeling \(p^*(t|m)\) by the composition of log-normal distributions.

\begin{table*}[!ht]
   \vskip -0.1in
   \caption{Mark prediction performance to evaluate \textit{ours} against existing \acrshort*{mtpp} models, measured by macro-F1/micro-F1. The bold are the best values.}
   \centering
   \begin{small}
   \setlength{\tabcolsep}{2pt}
   \resizebox{\linewidth}{!}{
   \begin{tabular}{llcccc}
       \toprule
                                & & Retweet & SO & Taobao & USearthquake \\
       \midrule
       \multirow{3}{*}{\makecell{{ours}}}   & \(\mathrm{M}\) & \textbf{0.4750\tiny{$\pm$0.0033}}/\textbf{\meanstd{0.4394}{0.0093}} & \textbf{0.1776\tiny{$\pm$0.0030}}/\textbf{\meanstd{0.6376}{0.0026}} & \textbf{0.4190\tiny{$\pm$0.0104}}/\textbf{\meanstd{0.7499}{0.0151}} & \textbf{0.1382\tiny{$\pm$0.0071}}/\textbf{\meanstd{0.3189}{0.0125}} \\
       & \(\mathrm{M}_r\) & \textbf{0.2010\tiny{$\pm$0.0082}}/\textbf{\meanstd{0.2010}{0.0082}} & \textbf{0.1476\tiny{$\pm$0.0041}}/\textbf{\meanstd{0.4530}{0.0026}} & \textbf{0.3987\tiny{$\pm$0.0108}}/\textbf{\meanstd{0.7558}{0.0185}} & \textbf{0.0339\tiny{$\pm$0.0051}}/\textbf{\meanstd{0.1111}{0.0098}} \\
       & \(\mathrm{M}_f\) & {0.6120\tiny{$\pm$0.0013}}/{\meanstd{0.9612}{0.0021}} & \textbf{0.2795\tiny{$\pm$0.0014}}/\textbf{\meanstd{0.8974}{0.0042}} & \textbf{0.7441\tiny{$\pm$0.0060}}/\textbf{\meanstd{0.7441}{0.0060}} & {0.2773\tiny{$\pm$0.0215}}/{\meanstd{0.9181}{0.0102}} \\

       \midrule
       \multirow{3}{*}{FullyNN}           & \(\mathrm{M}\) & {0.2190\tiny{$\pm$0.0000}}/\meanstd{0.0000}{0.0000} & {0.0054\tiny{$\pm$0.0000}}/\meanstd{0.0000}{0.0000} & {0.0094\tiny{$\pm$0.0000}}/\meanstd{0.0000}{0.0000} & {0.0914\tiny{$\pm$0.0000}}/\meanstd{0.0000}{0.0000} \\
       & \(\mathrm{M}_r\)  & {0.0000\tiny{$\pm$0.0000}}/\meanstd{0.0000}{0.0000} & {0.0000\tiny{$\pm$0.0000}}/\meanstd{0.0000}{0.0000} & {0.0100\tiny{$\pm$0.0000}}/\meanstd{0.7209}{0.0000} & {0.0000\tiny{$\pm$0.0000}}/\meanstd{0.0000}{0.0000} \\
       & \(\mathrm{M}_f\) & {0.3284\tiny{$\pm$0.0000}}/\meanstd{0.9768}{0.0000} & {0.0236\tiny{$\pm$0.0000}}/\meanstd{0.9155}{0.0000} & {0.0000\tiny{$\pm$0.0000}}/\meanstd{0.0000}{0.0000} & {0.2134\tiny{$\pm$0.0000}}/\textbf{\meanstd{0.9457}{0.0000}} \\

       \midrule
       \multirow{3}{*}{SAHP}                  & \(\mathrm{M}\) & {0.4211\tiny{$\pm$0.0050}}/\meanstd{0.1540}{0.0480} & {0.1134\tiny{$\pm$0.0027}}/\meanstd{0.5665}{0.0059} & {0.0616\tiny{$\pm$0.0327}}/\meanstd{0.1650}{0.1574} & {0.0962\tiny{$\pm$0.0005}}/\meanstd{0.1237}{0.0060} \\
       & \(\mathrm{M}_r\) & {0.0266\tiny{$\pm$0.0135}}/\meanstd{0.0266}{0.0135} & {0.0863\tiny{$\pm$0.0032}}/\meanstd{0.3500}{0.0071} & {0.0269\tiny{$\pm$0.0341}}/\meanstd{0.0825}{0.1009} & {0.0037\tiny{$\pm$0.0010}}/\meanstd{0.0162}{0.0015} \\
       & \(\mathrm{M}_f\) & {0.6183\tiny{$\pm$0.0010}}/\meanstd{0.9769}{0.0001} & {0.2054\tiny{$\pm$0.0011}}/\meanstd{0.9170}{0.0005} & {0.6166\tiny{$\pm$0.0112}}/\meanstd{0.6166}{0.0112} & {0.2196\tiny{$\pm$0.0016}}/\meanstd{0.9451}{0.0002} \\

       \midrule
       \multirow{3}{*}{THP}            & \(\mathrm{M}\) & {0.2238\tiny{$\pm$0.0068}}/\meanstd{0.0000}{0.0000} & {0.0859\tiny{$\pm$0.0204}}/\meanstd{0.3984}{0.1867} & {0.0069\tiny{$\pm$0.0035}}/\meanstd{0.0000}{0.0000} & {0.0921\tiny{$\pm$0.0003}}/\meanstd{0.0000}{0.0000} \\
       & \(\mathrm{M}_r\) & {0.0000\tiny{$\pm$0.0000}}/\meanstd{0.0000}{0.0000} & {0.0519\tiny{$\pm$0.0270}}/\meanstd{0.2120}{0.1360} & {0.0074\tiny{$\pm$0.0037}}/\meanstd{0.7208}{0.0001} & {0.0000\tiny{$\pm$0.0000}}/\meanstd{0.0004}{0.0006} \\
       & \(\mathrm{M}_f\) & {0.3357\tiny{$\pm$0.0102}}/\meanstd{0.9768}{0.0000} & {0.2015\tiny{$\pm$0.0025}}/\meanstd{0.9140}{0.0012} & {0.0000\tiny{$\pm$0.0000}}/\meanstd{0.0000}{0.0000} & {0.2149\tiny{$\pm$0.0008}}/\textbf{\meanstd{0.9457}{0.0008}} \\

       \midrule
       \multirow{3}{*}{AttNHP}            & \(\mathrm{M}\) & {0.4100\tiny{$\pm$0.0049}}/\meanstd{0.1901}{0.0143} & {0.0594\tiny{$\pm$0.0037}}/\meanstd{0.4548}{0.0148} & {0.2930\tiny{$\pm$0.0353}}/\meanstd{0.6359}{0.0415} & {0.1306\tiny{$\pm$0.0041}}/\meanstd{0.0809}{0.0460} \\
       & \(\mathrm{M}_r\) & {0.0373\tiny{$\pm$0.0056}}/\meanstd{0.0373}{0.0056} & {0.0188\tiny{$\pm$0.0000}}/\meanstd{0.2476}{0.0029} & {0.2682\tiny{$\pm$0.0363}}/\meanstd{0.5868}{0.0604} & {0.0012\tiny{$\pm$0.0007}}/\meanstd{0.0092}{0.0079} \\
       & \(\mathrm{M}_f\) & {0.5963\tiny{$\pm$0.0046}}/\meanstd{0.9747}{0.0007} & {0.1972\tiny{$\pm$0.0164}}/\meanstd{0.8372}{0.0643} & {0.6901\tiny{$\pm$0.0189}}/\meanstd{0.6901}{0.0189} & \textbf{0.3031\tiny{$\pm$0.0086}}/\meanstd{0.9434}{0.0012} \\

       \midrule
       \multirow{3}{*}{Marked-LNM} & \(\mathrm{M}\) & {0.4216\tiny{$\pm$0.0021}}/\meanstd{0.1565}{0.0129} & {0.1323\tiny{$\pm$0.0009}}/\meanstd{0.5995}{0.0038} & {0.0911\tiny{$\pm$0.0551}}/\meanstd{0.6658}{0.0615} & {0.1056\tiny{$\pm$0.0048}}/\meanstd{0.1130}{0.0027} \\
       & \(\mathrm{M}_r\) & {0.0252\tiny{$\pm$0.0041}}/\meanstd{0.0252}{0.0041} & {0.1119\tiny{$\pm$0.0001}}/\meanstd{0.3940}{0.0055} & {0.0547\tiny{$\pm$0.0577}}/\meanstd{0.6278}{0.0884} & {0.0063\tiny{$\pm$0.0072}}/\meanstd{0.0135}{0.0006} \\
       & \(\mathrm{M}_f\) & \textbf{0.6198\tiny{$\pm$0.0010}}/\textbf{\meanstd{0.9769}{0.0000}} & {0.2016\tiny{$\pm$0.0004}}/\meanstd{0.9123}{0.0011} & {0.7077\tiny{$\pm$0.0308}}/\meanstd{0.7077}{0.0308} & {0.2380\tiny{$\pm$0.0030}}/\meanstd{0.9451}{0.0001} \\
   \bottomrule
   \end{tabular}
   }
   \end{small}
   \label{tab:IFIB_real_world_f1}
\end{table*}

\section{Related Work}\label{sec:relatedwork}
\textbf{Marked Temporal Point Process}
Many \acrshort*{mtpp} studies specify a separate \acrfull*{cif} \(\lambda^*(m, t)\) for each categorical mark $m$, based on which \(p^*(m, t)\) can be formulated \citep{daleyIntroductionTheoryPoint2003, mei_neural_2017, zuo_transformer_2020, zhang_self-attentive_2020, enguehard_neural_2020, mei_transformer_2021, panosDecomposableTransformerPoint2024}. A more sophisticated intensity function \citep{mei_neural_2017, zuo_transformer_2020, zhang_self-attentive_2020, mei_transformer_2021} can better capture the system dynamics but will require approximating the integral of \(\lambda^*(m, t)\) using a numerical method such as Monte Carlo. Recurrent Marked Temporal Point Process(RMTPP) \citep{du_recurrent_2016} eludes numerical integral approximation as the \acrshort*{cif} and its integral have a closed form, which makes the log-likelihood easy to compute. Recent studies move away from directly modeling \acrshort*{cif}. \citet{shchur_intensity-free_2020} proposed an intensity-free solution, called LogNormMix, to infer the density function $p^*(t)$ from a simple distribution such as the mixture of log-normal distributions. \citet{omi_fully_2019} proposed FullyNN to model the integral of \acrshort*{cif} using a neural network where \acrshort*{cif} can be derived by differentiation, an operation computationally much easier compared with integration. All \acrshort*{mtpp} studies discussed so far predict the time of the next event first and then predict the mark. Recently, \citet{waghmare_modeling_2022} proposes to model \(p^*(m)\) using a classifier to predict the mark of the next event and modeling \(p^*(t|m)\) to predict the time of the event based on LogNormMix.

Recently, \citet{yuan_spatio-temporal_2023} used a \acrfull*{ddpm} to predict the next event in the spatio-temporal point process. \citet{ludke_add_2023} developed Add-and-thin, a method for modifying event sequences sampled from a Poisson process to match a target distribution by adding or removing events. However, the mark of the spatio-temporal point process is continuous instead of discrete, and Add-and-thin is a temporal point process (TPP) model that does not consider marks. Therefore, these two approaches are out of the scope of our research.

\textbf{Imbalanced Data Handling}
The techniques for handling imbalanced data, including data-level, algorithm-level, and classifier-level approaches, are designed mainly for improving imbalanced classification tasks. The data-level approach is resampling the training set, including undersampling and oversampling \citep{aminianChebyshevApproachesImbalanced2021}. Most existing resampling methods are based on the Synthetic Minority Over-sampling Technique (SMOTE) algorithm \citep{fernandezSMOTELearningImbalanced2018, bernardoCSMOTEContinuousSynthetic2020, bernardoVFCSMOTEVeryFast2021}. One benefit of data-level approaches is that they can cooperate with any classifiers. In contrast, algorithm-level approaches are more classifier-specific, such as cost-sensitive methods \citep{10.1007/978-3-319-71246-8_31, loezer2020cost, CHEN2024111272}.
The classifier-level method is also known as \textit{thresholding} (or \textit{post-scaling}) which learns thresholds to tune the obtained class probability \citep{lawrenceNeuralNetworkClassification2012, BUDA2018249, chanApplicationDecisionRules2019, tianPosteriorRecalibrationImbalanced2020}.
The effectiveness of resampling the train set is determined by the resampling ratio, but there is no easy way to figure it out for different classes on different datasets. The cost-sensitive approaches require domain knowledge regarding the importance of different marks to set the cost, but this is not always available \citep{johnsonSurveyDeepLearning2019}. To have a solution with minimum external knowledge and assumptions, this study adopts thresholding.

\section{Conclusion and Limitation}
\textbf{Conclusion}\quad
It is challenging for existing \acrshort*{mtpp} methods to accurately predict events of rare marks when the distribution of event marks is highly imbalanced. This is unacceptable in many applications if the rare mark is critical such as major earthquakes. This study introduces the first solution to address mark imbalance in \acrshort*{mtpp}. Instead of predicting mark based on mark probability directly as in existing studies, we learn thresholds to tune the mark probability normalized by the prior probability to optimize mark prediction. To achieve this goal, this study develops a strategy to predict mark first and then the time by integrating two improper integrations into one and proposing a novel \acrfull*{model} to approximate the unified improper integration to support time sampling and estimation of mark probability, rather than using computationally expensive numerical improper integration. Extensive experiments on real-world datasets demonstrate the superior performance of our solution against various baselines in the next event mark and time prediction.

\textbf{Limitation}\quad
As the first effort to address the mark imbalance for \acrshort*{mtpp}, this study verifies the effectiveness of thresholding, but does not investigate (i) the opportunity to extend the thresholding method to incorporate domain knowledge, such as the importance of rare marks, and (ii) the effectiveness of resampling and cost-sensitive approaches in this situation.


\section*{Broader Impact}
This paper presents work whose goal is to advance the field of Machine Learning. Specifically, we want to reveal the mark imbalance to the MTPP community and propose a relatively simple solution to inspire the development of more bias-aware MTPP approaches. There are many potential societal consequences of our work, none of which we feel must be specifically highlighted here.

\section*{Acknowledgement}
This research is supported in part by the Australian Research Council (ARC) Discovery Projects DP200101441 and DP210100743.

\newpage
\bibliography{reference.bib}
\bibliographystyle{abbrvnat}

\newpage
\section*{NeurIPS Paper Checklist}
\begin{enumerate}

    \item {\bf Claims}
        \item[] Question: Do the main claims made in the abstract and introduction accurately reflect the paper's contributions and scope?
        \item[] Answer: \answerYes{} 
        \item[] Justification: The contribution of this paper is first realizing the impact of mark imbalance to event predictions of MTPPs then devising an MTPP model and additional techniques to tackle this issue. They are reflected in our abstract and introduction.
    
    \item {\bf Limitations}
        \item[] Question: Does the paper discuss the limitations of the work performed by the authors?
        \item[] Answer: \answerYes{} 
        \item[] Justification: We have discussed the limit of our work in the main paper.
    
    \item {\bf Theory assumptions and proofs}
        \item[] Question: For each theoretical result, does the paper provide the full set of assumptions and a complete (and correct) proof?
        \item[] Answer: \answerYes{} 
        \item[] Justification: We provide the complete proof of how to obtain \(p^*(m, t)\) in Appendix A and \(\Gamma^*(m, t)\) in the main paper.
    
        \item {\bf Experimental result reproducibility}
        \item[] Question: Does the paper fully disclose all the information needed to reproduce the main experimental results of the paper to the extent that it affects the main claims and/or conclusions of the paper (regardless of whether the code and data are provided or not)?
        \item[] Answer: \answerYes{} 
        \item[] Justification: We have provided information including datasets, hyperparameters, technical details for reproducibility.

    \item {\bf Open access to data and code}
        \item[] Question: Does the paper provide open access to the data and code, with sufficient instructions to faithfully reproduce the main experimental results, as described in supplemental material?
        \item[] Answer: \answerYes{} 
        \item[] Justification: A sample code with running scripts and instructions are available in our supplementary material. They will be publically available upon acceptance.

    \item {\bf Experimental setting/details}
        \item[] Question: Does the paper specify all the training and test details (e.g., data splits, hyperparameters, how they were chosen, type of optimizer, etc.) necessary to understand the results?
        \item[] Answer: \answerYes{} 
        \item[] Justification: All training details are covered in the Appendix.
    
    \item {\bf Experiment statistical significance}
        \item[] Question: Does the paper report error bars suitably and correctly defined or other appropriate information about the statistical significance of the experiments?
        \item[] Answer: \answerYes{} 
        \item[] Justification: We report the 1-sigma error on all experiment results and state this fact at the beginning of the experiment section.
    
    \item {\bf Experiments compute resources}
        \item[] Question: For each experiment, does the paper provide sufficient information on the computer resources (type of compute workers, memory, time of execution) needed to reproduce the experiments?
        \item[] Answer: \answerYes{} 
        \item[] Justification: We expressed that we use A100 GPUs to run our experiments in the Appendix.
        
    \item {\bf Code of ethics}
        \item[] Question: Does the research conducted in the paper conform, in every respect, with the NeurIPS Code of Ethics \url{https://neurips.cc/public/EthicsGuidelines}?
        \item[] Answer: \answerYes{} 
        \item[] Justification: This work does not involve human subjects or participants. The datasets are publically available without copyright requirements. We also can not find any social harm, such as safety issues, security issues, discrimination, etc., that our approach may cause. 

    \item {\bf Broader impacts}
        \item[] Question: Does the paper discuss both potential positive societal impacts and negative societal impacts of the work performed?
        \item[] Answer: \answerYes{} 
        \item[] Justification: We have a broader impact discussing the potential impact of our approach at the end of the Appendix.
        
    \item {\bf Safeguards}
        \item[] Question: Does the paper describe safeguards that have been put in place for responsible release of data or models that have a high risk for misuse (e.g., pretrained language models, image generators, or scraped datasets)?
        \item[] Answer: \answerNA{} 
        \item[] Justification: Our paper poses no such risks of misuse.
    
    \item {\bf Licenses for existing assets}
        \item[] Question: Are the creators or original owners of assets (e.g., code, data, models), used in the paper, properly credited and are the license and terms of use explicitly mentioned and properly respected?
        \item[] Answer: \answerYes{} 
        \item[] Justification: We have presented the creaters and owners of all assets we used in our paper.
    
    \item {\bf New assets}
        \item[] Question: Are new assets introduced in the paper well documented and is the documentation provided alongside the assets?
        \item[] Answer: \answerYes{} 
        \item[] Justification: These documents are available alongside the code.
    
    \item {\bf Crowdsourcing and research with human subjects}
        \item[] Question: For crowdsourcing experiments and research with human subjects, does the paper include the full text of instructions given to participants and screenshots, if applicable, as well as details about compensation (if any)? 
        \item[] Answer: \answerNA{} 
        \item[] Justification: This paper does not involve crowdsourcing nor research with human subjects.
    
    \item {\bf Institutional review board (IRB) approvals or equivalent for research with human subjects}
        \item[] Question: Does the paper describe potential risks incurred by study participants, whether such risks were disclosed to the subjects, and whether Institutional Review Board (IRB) approvals (or an equivalent approval/review based on the requirements of your country or institution) were obtained?
        \item[] Answer: \answerNA{} 
        \item[] Justification: This paper does not involve crowdsourcing nor research with human subjects.
    
    \item {\bf Declaration of LLM usage}
        \item[] Question: Does the paper describe the usage of LLMs if it is an important, original, or non-standard component of the core methods in this research? Note that if the LLM is used only for writing, editing, or formatting purposes and does not impact the core methodology, scientific rigorousness, or originality of the research, declaration is not required.
        \item[] Answer: \answerNA{} 
        \item[] Justification: The LLM is used only for writing, editing, or formatting purposes and does not impact the core methodology, scientific rigorousness, or originality of the research.
    
    \end{enumerate}

\newpage

\appendix
\section{The Conditional Joint PDF}
\label{appendix:mtpp_pdf}
This study concerns events with categorical marks. For mark \(m\), we define a conditional intensity function \(\lambda^*(m, t)\):
\begin{equation}
\begin{aligned}
\label{eqn:lambda_i}
    &\lambda^*(m = k_i, t) = \lambda(m = k_i, t|\mathcal{H}_t) \\
    = & \lim_{\Delta t \rightarrow 0}{\frac{P(m = k_i, t \in [t, t + \Delta t)|\mathcal{H}_{t-})}{\Delta t}} \\
    = & \lim_{\Delta t \rightarrow 0}{\frac{p(m = k_i, t \in [t, t + \Delta t)|\mathcal{H}_{t_l}) \Delta t}{P(\forall j \in \mathbb{N}^+, t_j \notin (t_l, t)|\mathcal{H}_{t_l}) \Delta t}} \\
    = & \lim_{\Delta t \rightarrow 0}{\frac{p(m = k_i, t \in [t, t + \Delta t)|\mathcal{H}_{t_l})}{P(\forall j \in \mathbb{N}^+, t_j \notin (t_l, t)|\mathcal{H}_{t_l})}} \\
    = & \frac{p(m = k_i, t \in [t, t + dt)|\mathcal{H}_{t_l})}{P(\forall j \in \mathbb{N}^+, t_j \notin (t_l, t)|\mathcal{H}_{t_l})}
\end{aligned}
\end{equation}
where \(\mathcal{H}_{t_l}\) is the history up to (including) the most recent event, \(\mathcal{H}_{t-}\) is the history up to (excluding) the current time, \(P(\forall j \in \mathbb{N}^+, t_j \notin (t_l, t)|\mathcal{H}_{t_l})\) represents the probability that no event is observed in time interval \((t_l, t)\) given \(\mathcal{H}_{t_l}\).

We denote \(P^{'}_m((t_1,t_2)|\mathcal{H}_{t_l})\) for the conditional probability that an event \(m\) happens in $(t_1,t_2)$.
Following the definition of simple TPP that at most one event happens at every timestamp \(t\), the probability that no event occurs in \((t_l, t)\) is:
\begin{equation}
\begin{aligned}\label{eqn:noevent}
& P(\forall j \in \mathbb{N}^+, t_j \notin (t_l, t)|\mathcal{H}_{t_l}) \\ 
=& 1 - \sum_{m\in \mathrm{M}}{P^{'}_m((t_l,t)|\mathcal{H}_{t_l})\prod_{n\in \mathrm{M}, n \neq m}({1 - P^{'}_n((t_l,t)|\mathcal{H}_{t_l})})} \\
=& 1 - \sum_{m\in \mathrm{M}}{\frac{P^{'}_m((t_l,t)|\mathcal{H}_{t_l})}{1-P^{'}_m((t_l,t)|\mathcal{H}_{t_l})}}\prod_{n\in M}({1-P^{'}_n((t_l,t)|\mathcal{H}_{t_l})}) \\
=& 1 - \sum_{m\in \mathrm{M}}{F(m,t|\mathcal{H}_{t_l})} = 1 - \sum_{m\in \mathrm{M}}{F^*(m,t)} 
\end{aligned}
\end{equation}
where
\begin{equation}\label{eqn:P_mt}
F^*(m,t) = \frac{P^{'}_m((t_l,t)|\mathcal{H}_{t_l})}{1-P^{'}_m((t_l,t)|\mathcal{H}_{t_l})}\prod_{n\in \mathrm{M}}({1 - P^{'}_n((t_l, t)|\mathcal{H}_{t_l}))}
\end{equation}
The conditional joint PDF that the next event is $m$ and occurs in $[t, t+dt)$ is:
\begin{subequations}
\begin{align}\label{eqn:pnext}
    & p(m = k_i, t \in [t, t + \Delta t)|\mathcal{H}_{t_l}) = \frac{dF^*(m = k_i,t)}{dt} \\ 
    & \int_{t_l}^{t}{p(m = k_i, t \in [t, t + \Delta t)|\mathcal{H}_{t_l})d\tau} = F^*(m = k_i, t)
\end{align}
\end{subequations}

In this study, $p^*(m,t)$, shorthand of \(p(m, t|\mathcal{H}_{t_l})\), is the formal representation of $p(m = k_i, t \in [t, t + \Delta t)|\mathcal{H}_{t_l})$. Note $F^*(m, t)$ in \cref{eqn:P_mt} is the probability that only one event happens in interval \([t, t + dt)\) and the mark is $m$. This ensures the \acrshort*{mtpp} represented by $p^*(m,t)$ is simple. By integrating \cref{eqn:pnext} and \cref{eqn:noevent} in \cref{eqn:lambda_i}, we have  
\begin{equation}
  p^*(m, t) = \lambda^*(m, t)(1 - \sum_{w\in \mathrm{M}}{F^*(w,t)})
\end{equation}
where $\sum_{w\in \mathrm{M}}{F^*(w,t)}$ is calculated from the sum of \cref{eqn:lambda_i} over marker \(m\):
\begin{equation}
\label{eqn:mtpp_sum_P}
    \sum_{w\in \mathrm{M}}{F^*(w, t)} = 1 - \exp(-\int_{t_l}^{t}{\sum_{n\in \mathrm{M}}{\lambda^*(n, \tau)}d\tau})
\end{equation}
Then, we solve $p^*(m, t)$:
\begin{equation}
p^*(m, t) = \lambda^*(m, t)\exp(-\int_{t_l}^{t}{\sum_{n\in \mathrm{M}}{\lambda^*(n, \tau)}d\tau})
\end{equation}
which is equivalent with \cref{eqn:mtpp}. 

\section{Proof of the Proposition 3.1}
\label{appendix:ifnmtpp_correct}
\begin{proof}
The gradient of IFNMTPP is \(-p^*(m, t)\), the function it learns must take the form:

\begin{equation}
\mathrm{IFNMTPP}(m, t) = -\int_{C_1}^{t}{p^*(m, \tau)d\tau} + C_2
\label{seq:what_we_learns}
\end{equation}
where \(C_1\) and \(C_2\) are two constants. 

From IFNMTPP architecture, the Integral Estimation Module (IEM) consists of multiple fully connected layers with non-negative weights, and monotonic increasing and unbounded activation functions. Then, it ends with a monotonic decreasing function \(\sigma(x) = \frac{1}{1 + e^x}\) (as illustrated in \cref{fig:IFIB}). Since \(\lim_{x \to +\infty} \sigma(x) = 0\), we have:
\begin{equation}
\lim_{t\rightarrow+\infty}{\mathrm{IFNMTPP}(m, t)} = \lim_{t\rightarrow+\infty}{-\int_{C_1}^{t}{p^*(m, \tau)d\tau} + C_2} = 0
\end{equation}
Substituting this into the earlier equation, we obtain:
\begin{equation}
C_2 = \lim_{t\rightarrow+\infty}{\int_{C_1}^{t}{p^*(m, \tau)d\tau}} = \int_{C_1}^{+\infty}{p^*(m, \tau)d\tau}
\end{equation}
Substituting \(C_2\) back in \cref{seq:what_we_learns}:
\begin{equation}
\mathrm{IFNMTPP}(m, t) = \int_{C_1}^{+\infty}{p^*(m, \tau)d\tau}-\int_{C_1}^{t}{p^*(m, \tau)d\tau} = \int_{t}^{+\infty}{p^*(m, \tau)d\tau} = \Gamma^*(m, t)
\end{equation}
Thus, the output of IFNMTPP is \(\Gamma^*(m, t)\) when its gradient is \(-p^*(m, t)\).
\end{proof}

\section{Technical Details}
\subsection{Technical Details about IFNMTPP}
\label{app:det_ifnmtpp}
In \cref{appendix:ifnmtpp_correct}, we show that IFNMTPP models \(\Gamma^*(m, t)\) when the activation function in the Integral Estimation Module (IEM) is monotonic increasing and unbounded. However, we select \(\mathtt{tanh}\) as the activation function for training stability in the implementation. \(\mathtt{tanh}\) is monotonic but bounded, so \(\lim_{x \to +\infty} \mathrm{IFNMTPP}(m, t) = C > 0\), making the implemented IFNMTPP slightly inaccurate. To mitigate this issue, we subtract the original output from the implemented IFNMTPP with \(C\). The pseudo code is below:
\begin{lstlisting}[language=Python]

for layer_idx, layer in enumerate(self.mlp):
    # Hidden status at t for calculating \Gamma^*(m, t)
    output = layer(output)
    output = self.layer_activation(output)

    # Hidden status at t_l for calculating \Gamma^*(m, t_l)
    output_zero = layer(output_zero)
    output_zero = self.layer_activation(output_zero)

    if layer_idx == 0:
        # Hidden status at infinity for calculating \Gamma^*(m, +\infty) a.k.a. C.
        output_max = torch.ones_like(output) * self.tanh_parameter
    else:
        output_max = layer(output_max)
        output_max = self.layer_activation(output_max)

probability_integral_from_t_to_inf = self.nonneg_integral(-self.aggregate(output))

probability_integral_from_tl_to_inf = self.nonneg_integral(-self.aggregate(output_zero))

probability_integral_minimal = self.nonneg_integral(-self.aggregate(output_max))

# Shift the output with C when required.
if self.removes_tail:
    regularized_probability_integral_from_t_to_inf = (probability_integral_from_t_to_inf - probability_integral_minimal)

    regularized_probability_integral_from_tl_to_inf = (probability_integral_from_tl_to_inf - probability_integral_minimal) + self.epsilon

else:
    regularized_probability_integral_from_t_to_inf = probability_integral_from_t_to_inf

    regularized_probability_integral_from_tl_to_inf = probability_integral_from_tl_to_inf + self.epsilon
\end{lstlisting}

\subsection{Technical Details about Thresholding}
\label{app:thresholding}
We use the classic threshold-tuning method \citep{lawrenceNeuralNetworkClassification2012, BUDA2018249} to obtain the optimal $\epsilon_m$. Specifically, the method obtains optimal $\epsilon_m$ by taking three steps for each mark $m$. In step 1, we draw the precision-recall curve of $m$. This curve shows us the precision and recall across all possible thresholds. In step 2, since our target is to maximize the F1 score, which is the harmonic mean of the precision and recall, we compute the F1 score for all possible threshold using the precision and recall obtained in the first step. In step 3, the threshold that yields the maximum F1 value is the $\epsilon_m$. The pseudo code is below:

\begin{lstlisting}[language=Python]
# For each mark $m \in M$ 
for i in range(num_events): 

    # Step 1: draw the precision-recall curve. 
    precision[i], recall[i], thresholds = precision_recall_curve((training_results_events_next == i).astype(int), scaled_training_results_pm[:, i])
  
    # Step 2: Calculate the F1 score across all possible threshold. 
    f1s = (2 * precision[i] * recall[i]) / (precision[i] + recall[i]) 
    f1s = np.nan_to_num(f1s) 

    # Step 3: Pick the threshold that yields the maximal F1 value. 
    ix = np.argmax(f1s) 
    f1[i] = f1s[ix] 
    threshold.append(thresholds[ix]) 
\end{lstlisting}

Once $\epsilon_m$ is known for each mark $m \in M$, we predict the next mark as $m$ if $r_m-\epsilon_m$ is maximum. Why? because such $m$ will lead to a higher F1 value compared to other marks.

Please note that we did not train a machine learning model like a neural network to obtain the optimal $\epsilon_m$. Conceptually, one can predict $m$ by selecting $m$ with the maximum $r_m-\epsilon_m$ (i.e., $argmax$) and use the prediction loss to update the parameters of the model producing $\epsilon_m$. However, $argmax$ is non-differentiable so that backpropagation is not allowed to update model parameters. This is why we did not use this method.

\section{Experiment Settings}
\label{app:hyperparameters}
\subsection{Real-world Datasets}
\label{app:realworld_generation}
We use the following four datasets to evaluate the performance of \acrshort*{model}.
\begin{itemize}
    \item \textit{Retweet dataset}\citep{zhao_seismic_2015} records when users Retweet a particular message on Twitter. This dataset distinguishes all users into three different types: (1) normal user, whose followers count is lower than the median, (2) influence user, whose followers count is higher than the median but lower than the $95$th percentile, (3) famous user, whose followers count is higher than the $95$th percentile. About 2 million Retweets are recorded, and the average sequence length is 108.This dataset is released under the Apache-2.0 license\citep{xue_easytpp_2023}.
    \item \textit{StackOverflow dataset}(SO)\citep{Leskovec2014SNAPD} was collected from Stackoverflow\footnote{https://StackOverflow.com/}, a popular question-answering website about various topics. Users providing decent answers will receive different badges as rewards. This dataset collects the timestamps when people obtain 22 badges from the website, and the average sequence length is 72.This dataset is released under the Apache-2.0 license\citep{xue_easytpp_2023}.
    \item \textit{Taobao}\citep{taobao} records users' interactions on Taobao, an online shopping website from China. These actions include user clicking and buying online items, viewing reviews and comments, or searching for items. The average length of sequences in this dataset is 58, and 17 different marks are available. This dataset is released under the Apache-2.0 license\citep{xue_easytpp_2023}.
    \item \textit{USearthquake}\citep{xue_easytpp_2023} records all earthquakes happened in the continental US from USGS\footnote{http://earthquake.usgs.gov/earthquakes/eqarchives/year/eqstats.php}. This dataset has 7 marks, referring to earthquakes with magnitude 2.0 to 2.9, 3.0 to 3.9, 4.0 to 4.9, 5.0 to 5.9, 6.0 to 6.9, 7.0 to 7.9, or 8 and higher. The average sequence length is 16.This dataset is released under the Apache-2.0 license\citep{xue_easytpp_2023}.
\end{itemize}

\subsection{Synthetic Datasets}
All synthetic datasets are generated so we do not have any licenses information for them. The code to generate all synthetic datasets comes from the codebase of \citep{omi_fully_2019} at \url{https://github.com/omitakahiro/NeuralNetworkPointProcess} which is publicly accessible without any licenses.
\label{app:synthetic_generation}
\begin{itemize}
\item 
\textit{Hawkes process dataset Hawkes\_1} was generated utilising Hawkes process: 
\begin{equation}
\lambda^{*}(t) = \mu_0 + \sum_{t_i < t}{a\exp(-b(t - t_i))}    
\end{equation}
where \(\mu = 0.2\), \(a = 0.8\), and \(b = 1.0\).
     
\item 
\textit{Hawkes process dataset Hawkes\_2} was generated utilising Hawkes process: 
\begin{equation}
\lambda^{*}(t) = \mu_0 + \sum_{t_i < t}{a_1\exp(-b_1(t - t_i)) + a_2\exp(-b_2(t - t_i))}    
\end{equation}
where \(\mu = 0.2\), \(a_1 = a_2 = 0.4\), \(b_1 = 1.0 \), and \(b_2 = 20\).

\item 
\textit{Homogeneous Poisson process dataset} was generated using the Homogeneous Poisson process where the conditional intensity function \(\lambda^*(t)\) is constant over the entire timeline. This paper assumes \(\lambda^*(t) = 1\).

\item 
\textit{Self-correct process dataset} was generated using the temporal point process whose intensity significantly drops when an event happens. The definition of the conditional intensity function is \(\lambda^*(t) = \exp(\mu(t - t_i) - \alpha N)\) where \(N\) is the number of occurred events, and \(\mu\) and \(\alpha\) are fixed parameters. In our experiments, we set \(\alpha = \mu = 1\).

\item 
\textit{Stationary renewal process dataset} was generated using stationary renewal process, which directly defines the probability distribution over time \(p^*(t)\) as a log-normal distribution as shown in \cref{eqn:log_norm}.
\begin{equation}
\label{eqn:log_norm}
p^*(t|\sigma) = \frac{1}{\sigma t\sqrt{2\pi}}\exp(-\frac{\log^2(t)}{2\sigma^2})
\end{equation}
where \(\sigma\) is the standard deviation. Here, we set \(\sigma = 1\). With \cref{eqn:log_norm} and TPP's definition, one could solve the corresponding intensity function by Wolframalpha\footnote{https://www.wolframalpha.com}:
\begin{equation}
\label{eqn:intensity}
    \lambda^*(t) = \frac{-0.797885\exp(-0.5\log^2(t))}{-t + t \operatorname{erf}(0.707107\log(t))}
\end{equation}
where \(\operatorname{erf}(x) = \frac{2}{\sqrt{\pi}}\int_{0}^{x}{\exp(-t^2)dt}\).
\end{itemize}

These five synthetic distributions cooperate with a synthetic marking methods. This method generates discrete marks sampled from a uniform distribution. All synthetic datasets have 5 different marks.

\subsection{Metrics}\label{app:metric}
\subsubsection{Metrics for Synthetic Datasets}
\label{app:metric_syn}
For synthetic datasets, the real distribution $\hat{p}^*(m, t)$ is known. We can compare the generated $p^*(m, t)$ against the real one. Most papers report the relative NLL loss, that is, the average of the absolute difference between \(-\log\hat{p}^*(m, t)\) and \(-\log{p}^*(m, t)\) on the observed events(if markers are unavailable, \(-\log\hat{p}^*(t)\) and \(-\log{p}^*(t)\)\citep{omi_fully_2019, shchur_intensity-free_2020}). The lower relative NLL loss indicates a better performance. However, such a metric only evaluates performance at discrete events, which cannot gauge the overall discrepancy between $\hat{p}^*(m, t)$ and $p^*(m, t)$. So, this paper selects Spearman Coefficient \(\rho\) and \(L^1\) distance to measure the discrepancy between $\hat{p}^*(m, t)$ and $p^*(m, t)$ over time, while we also report the relative NLL loss for reference.

\textit{Spearman Coefficient} $\rho(X, Y)$ measures the relationship between two arbitrary value sequences, \(X\) and \(Y\), as defined by \cref{eqn:spearman}. If \(X\) and \(Y\) are more correlated, $\rho(X, Y)$ is higher; lower otherwise. Compared with the Pearson coefficient which is suitable if the relationship between \(X\) and \(Y\) is linear, Spearman coefficient could better deal with non-linear relationships. Because most probability distributions of TPP are non-linear, we select Spearman coefficient.
\begin{equation}
\label{eqn:spearman}
    \rho(X, Y) = \frac{\operatorname{Cov}(\operatorname{Rank}(X), \operatorname{Rank}(Y))}{\sigma_{X}\sigma_{Y}} \in [-1, 1]
\end{equation}
where \(\sigma_{X}\) and \(\sigma_{Y}\) are the standard deviations of the values in sequence \(X = \{x_1, x_2, \cdots, x_n\}\) and \(Y = \{y_1, y_2, \cdots, y_n\}\), respectively. We expect \(\rho\) between $\hat{p}^*(m, t)$ and $p^*(m, t)$ is close to 1.

\textit{\(L^1\) distance} measures how different two arbitrary functions are in interval \([a, b]\).
\begin{equation}
    L^1(f, g) = \int_{a}^{b}{|f(x) - g(x)|dx} \geqslant 0
\end{equation}
The smaller the \(L^1\) distance is, the more similar $f(x)$ and $g(x)$ are. When \(L^1(f, g) = 0\), \(f(x)\) almost equals to \(g(x)\) in interval \([a, b]\) for any \(f(x)\) and \(g(x)\), or \(f(x) = g(x)\) at every \(x \in [a, b]\) if both \(f(x)\) and \(g(x)\) are continuous.

\subsubsection{Metrics for Real-World Datasets - macro-F1 \& micro-F1}
\label{app:mi-ma-f1}
The macro-F1 value and micro-F1 value derives from the F1 value. F1 value has been widely used in almost all binary classification tasks because, compared with accuracy that might be fooled by false positives, F1 value takes accuracy and recall rate in its mind, where the model should correctly mark out positive samples for a better accuracy and negative samples for a better recall rate. The definition of F1 value is:
\begin{equation}
    \operatorname{F1} = \frac{2\times \operatorname{Acc} \times \operatorname{Recall}}{\operatorname{Acc} + \operatorname{Recall}}
    \label{eqn:f1_def}
\end{equation}
F1 value is only for the binary classification. Some researchers realise that a multi-class classification can be evaluated by decomposing the original classification task into multiple binary classification tasks and averaging every obtained F1 values. This is how macro-F1 is devised. The expression of macro-F1 is:
\begin{equation}
    \operatorname{macro-F1} = \frac{1}{|M|}\sum_{m = 1}^{|M|}{\operatorname{F1}_m}
\end{equation}
where \(\operatorname{F1}_m\) is the F1 value for marker \(m\). macro-F1 treats all classes equally, so it has been widely used in studies addressing class imbalance.

On the other hand, micro-F1 is a global average of F1 values. Specifically, micro-F1 computes the sum of true positives, false negatives, and false positives over all classes then use \cref{eqn:f1_def} to obtain the micro-F1. micro-F1 shows the overall performance regardless of the class.

If the mark prediction is based on \(p^*(m)\) like our solution, macro-F1 and micro-F1 are independent of time prediction by nature. For baselines where mark prediction is based on \(p^*(m|t)\), the mark involved in macro-F1 and micro-F1 is conditioned on the real time of the next event to ensure that macro-F1 and micro-F1 are independent of the time prediction. Specifically, for each real next event \((m = k_i, t^{\prime})\) in a test set, we compute macro-F1 and micro-F1 using the mark predicted from \(p(m|t^{\prime})\).

\mysubsubsection[MAE]{Metrics for Real-World Datasets - \acrlong*{mmae} (MAE)}
\label{app:mae}
The test dataset \(T\) contains a subset of real next events. We denote \(T_{m=k_i}\subset T\) as those real next events where the mark is $k_i\in \mathrm{M}$. The number of events in \(T_{m=k_i}\) is \(|T_{m=k_i}|\). For each real next event \((m=k_i,t)\in T_{m=k_i}\), we are interested in the evaluation of time prediction.
Consider all real next events in \(T_{m=k_i}\), \acrshort*{mmae}\(_{m=k_i}\) can be defined:
\begin{equation}
    \operatorname{MAE}_{m=k_i} = \frac{1}{|T_{m=k_i}|}\sum_{(m=k_i,t)\in T_{m=k_i}}|t-\bar{t}_{m=k_i}|
\end{equation}
The absolute difference \(|t-\bar{t}_{m=k_i}|\), between real time \(t\) and the predicted time \(\bar{t}_{m=k_i}\) for mark $k_i$, is the prediction error for the real next event \((m=k_i,t)\). Here, $k_i$ is not necessarily the predicted mark so that the time prediction evaluation is independent of mark prediction. \acrshort*{mmae}\(_{\mathrm{M}_{*}}\) is the geometric mean of \acrshort*{mmae}\(_{m=k_i}\) across all marks in \(\mathrm{M}_{*}\). \(\mathrm{M}_{*}\) can be \(\mathrm{M}\), \(\mathrm{M}_f\), or \(\mathrm{M}_r\):
\begin{equation}
    \operatorname{MAE}_{\mathrm{M}_{*}} = \sqrt[^{|\mathrm{M}_{*}|}]{\prod_{k_i \in \mathrm{M_{*}}}{\operatorname{MAE}_{m=k_i}}}
\end{equation}
where \(|\mathrm{M}_{*}|\) is the number of marks in \(\mathrm{M}_{*}\).

\subsection{Baselines}
\label{app:baseline}
\subsubsection{Group One}
The first group of baselines includes: (i) \textit{ours-w/o-thresholding}, which is the same as our method but the mark prediction is not optimized with thresholding. The mark prediction returns the mark with the highest mark probability as described in \cref{sec:pre}. The purpose is to evaluate the effectiveness of the thresholding method. (ii) \textit{time-mark-with-thresholding}, that uses \acrshort*{model} to predict the time of the next event first, and then predicts the mark with the same thresholding method as \textit{ours}. To do that, we predict time \(\bar{t}\) which is the mean of \(N\) samples from \(p^*(t) = \sum_{m \in \mathrm{M}}{p^*(m, t)}\) first, and then modify \(p^*(m)\) in \cref{eqn:get_ratio} to \(p^*(m|\bar{t})\) for mark prediction following the procedure described in \cref{sec:thresholding}. The purpose is to evaluate the necessity to predict mark first for handling mark imbalance with thresholding. (iii) \textit{time-mark-w/o-thresholding}, is same as time-mark-with-thresholding but mark prediction is not optimized with thresholding.

\subsubsection{Group Two}
The second group of baselines evaluates thresholding against resampling, another classic technique to address data imbalance, including \textit{undersampling} and \textit{oversampling}. For undersampling, we reduce the frequency of other marks to ensure that they have the same number of training events as the rarest mark. For oversampling, we increase the frequency of other marks so that they have the same number of training events as the most frequent mark. For a fair comparison, the backbone \acrshort*{mtpp} method is \acrshort*{model}. After training completes for both baselines, the mark with the highest probability is predicted as the next event mark.

\subsubsection{Group Three}
The third group of baselines includes existing \acrshort*{mtpp} methods. Since \acrshort*{mtpp} modeling has been well studied in the past decades, the state-of-the-art methods demonstrate comparable performance. Among them, this study selects the most popular ones as baselines. Four neural \acrshort*{mtpp} methods based on conditional intensity function (\acrshort*{cif}) are FullyNN \citep{omi_fully_2019}, THP \citep{zuo_transformer_2020}, SAHP \citep{zhang_self-attentive_2020}, and AttNHP \citep{mei_transformer_2021}. Besides these four, another baseline Marked-LNM \citep{waghmare_modeling_2022} models \(p^*(m)\) using a classifier to predict the mark of the next event and models \(p^*(t|m)\) using LogNormMix to predict the time of the event. 

\begin{itemize}
 \item \textit{Fully Neural Network(FullyNN)}\citep{omi_fully_2019} uses a neural network to estimate the integral of $\lambda^*(t)$ for the history embedding \(\mathbf{h}\) and inter-event time $t$. Then the density function is formulated to predict the time of the next event. We rewrote FullyNN in PyTorch\citep{NEURIPS2019_9015} based on the official implementation available at \url{https://github.com/omitakahiro/NeuralNetworkPointProcess}, which is publicly accessible without any license.
 
\item \textit{Transformer Hawkes Process(THP)}\citep{zuo_transformer_2020} uses a Transformer-based encoder to represent history as a hidden state \(\mathbf{h}\). The softplus-based intensity function and the density function are modelled to predict the time of next event. We reproduce this model in PyTorch based on the paper.

\item \textit{Self-Attentive Hawkes Process(SAHP)}\citep{zhang_self-attentive_2020} is based on the same intuition as Continuous-time LSTM(CTLSTM)\citep{mei_neural_2017}, which generalizes the classical Hawkes process by parameterizing its intensity function with recurrent neural networks. CTLSTM is an interpolated version of the standard LSTM, allowing us to generate outputs in a continuous-time domain. SAHP further improves performance by replacing LSTM with Transformers. Because the only difference between SAHP and CTLSTM is the history encoder, and SAHP has reported achieving better performance than CTLSTM, we only evaluate SAHP in this paper. We reproduce this model in PyTorch based on the paper.

\item \textit{AttNHP}\citep{mei_transformer_2021} is another Transformer-based \acrshort*{mtpp} model. Different from THP and SAHP, where Transformer only encodes history, and the distribution is extracted from history representations using another deep module, AttNHP merges these two modules into one by directly extracting the distribution from historical events using a Transformer. We use the code provided by the author at \url{https://github.com/yangalan123/anhp-andtt}.

\item \textit{Marked LogNormMix(Marked-LNM)}\citep{waghmare_modeling_2022} is an \acrshort*{mtpp} extension of the LogNormMix\citep{shchur_intensity-free_2020}. Marked-LNM also follows the \acrshort*{et} by modeling \(p^*(m)\) first, then using a composition of log Gaussian distribution to represent \(p^*(t|m)\). To the best of our knowledge, Marked-LNM is the only \acrshort*{mtpp} approach predicting the mark of the next event first and then predicting the time of the event. However, Marked-LNM limits the form of \(p^*(t|m)\) as the composition of log Gaussian distributions. This setting introduces inductive biases into the model, which could compromise the model prediction performance. We implement this model in PyTorch by modifying the official LogNormMix code at \url{https://github.com/shchur/ifl-tpp}. The official codes are released under the MIT license.
\end{itemize}

\subsection{Data Preprocessing}\label{apx:datapre}
We prepare synthetic and real-world datasets with normalization. For each dataset, normalization scales the time \(t\) of every event in each event sequence by the time mean \(\bar{t}\) of all events in all event sequences and standard deviation \(\sigma\), as shown in \cref{eqn:scale}:
\begin{equation}
    t_{scaled} = \frac{t - \bar{t}}{\sigma}
    \label{eqn:scale}
\end{equation}
Normalization is useful when the time is relatively large, such as in the Retweet dataset. \cref{tab:datasets} shows how normalization is applied on various datasets.
\begin{table}[!ht]
    \centering
    \caption{Data preprocessing.}
    \resizebox{\linewidth}{!}{
    \begin{tabular}{c|cccc|c}
    \toprule
         Dataset & Retweet & StackOverflow & Taobao & USearthquake & five synthetic datasets \\
    \hline
         Normalization & \checkmark & \checkmark & \checkmark & \checkmark & \ding{55} \\
    \bottomrule
    \end{tabular}}
    \vskip -0.1in
    \label{tab:datasets}
\end{table}

\begin{figure*}[ht]
\centering
    \begin{subfigure}{0.23\textwidth}
        \includegraphics[width=\textwidth]{dataset_distribution/retweet_mark_distribution.pdf}
        \caption{Retweet}
    \end{subfigure}
    \begin{subfigure}{0.23\textwidth}
        \includegraphics[width=\textwidth]{dataset_distribution/stackOverflow_mark_distribution.pdf}
        \caption{StackOverflow}
    \end{subfigure}
    \begin{subfigure}{0.23\textwidth}
        \includegraphics[width=\textwidth]{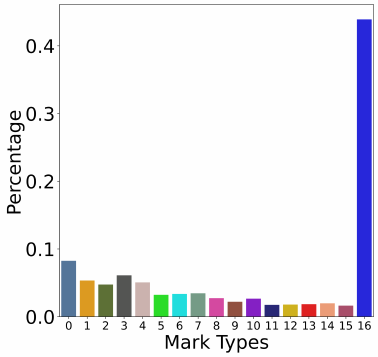}
        \caption{Taobao}
    \end{subfigure}
    \begin{subfigure}{0.23\textwidth}
        \includegraphics[width=\textwidth]{dataset_distribution/usearthquake_mark_distribution.pdf}
        \caption{USearthquake}
    \end{subfigure}
    \caption{The frequency distribution of marks in real-world datasets.}
    \label{fig:datset_mark_dist}
\end{figure*}

Our work focuses on predicting when the next event will happen provided a mark, especially a rare mark. 
For each dataset, we classify if one mark is rare or frequent. 
The percentages of marks in each dataset are presented in \cref{fig:datset_mark_dist}. \cref{tab:rare} shows which marks are classified as frequent and which are classified as rare. 

\begin{table}[!ht]
    \centering
    \caption{Rare marks and frequent marks.}
    \resizebox{\linewidth}{!}{
    \setlength{\tabcolsep}{2pt}
    \begin{tabular}{cccc}
    \toprule
         Dataset name & The number of marks & Rare Mark & Frequent Mark \\
    \midrule
         Retweet &  3 & [2] & [0, 1]\\
         StackOverflow & 22 & [1, 2, 6, 7, 9, 10, 11, 12, 13, 14, 15, 16, 17, 18, 19, 20, 21] & [0, 3, 4, 5, 8] \\
         Taobao & 17 & [0, 1, 2, 3, 4, 5, 6, 7, 8, 9, 10, 11, 12, 13, 14, 15] & [16]\\
         USearthquake & 7 & [3, 4, 5, 6] & [0, 1, 2]\\
    \bottomrule
    \end{tabular}}
    \vskip -0.1in
    \label{tab:rare}
\end{table}

\subsection{Model Training}
This section introduces the hyperparameter settings for all \acrshort*{mtpp} models used in this paper. The two values of ``Steps" refer to the number of warm-up steps and total training steps, respectively. ``BS" refers to batch size, and ``LR" refers to the learning rate. Unless otherwise specified, we repeatedly train a model 3 times with different random seeds and report the mean and standard deviation of the results. We conduct all experiments on an internal cluster. It includes Intel Xeon CPUs and NVIDIA A100-PCIE GPUs. All codes will be release upon acceptance under the MIT license.

For each mark $m$, we sample \(N\) times \(\{t^i\}_N^m\) from \(F^*(t|m)\) to predict the time of the next event on the condition that its mark is \(m\) by the inverse transform sampling:
\begin{equation}
\label{eqn:et_sample_recap}
    F^*(t^i|m) = u^i
\end{equation}
where \(u^i\) is a random sample from a uniform distribution. The common practice samples \(u_i\) from the standard uniform distribution \(u^i\sim \mathcal{U}(0, 1)\). \acrshort*{mtpp} allows \(t_i\) to go to positive infinity. When \(u^i\) is very close to 1, the time drawn from \cref{eqn:et_sample_recap} will be meaninglessly big and cause a negative impact to the accuracy of evaluation. To avoid this, we let \(u^i\sim \mathcal{U}(0, 0.9)\). We find this trick can significantly stabilize the sampling process.

\mysubsubsection[IFNMTPP Configurations]{\acrshort*{model} Configurations}
\cref{tab:IFIB_hyp} lists the hyperparameter settings for \acrshort*{model}. The three values of ``MS" (model structure) refer to the number of dimensions for history embedding \(\mathbf{h}\), the number of dimensions for \(\mathbf{v}_m\) and \(\mathbf{b}_m\)\footnote{\(\mathbf{v}_m\) and \(\mathbf{b}_m\) always have the same number of dimensions.}, and the number of non-negative fully-connected layers in the IEM module, respectively. 

\begin{table}[!ht]
    \centering
    \caption{Hyperparameter settings for \acrshort*{model}.}
    \label{tab:IFIB_hyp}
    \begin{footnotesize}
    \begin{tabular}{ccccc}
    \toprule
      Datasets & Steps & MS & BS & LR \\
    \midrule
      Retweet & [80,000, 400,000] & [32, 16, 4] & 32 & 0.002 \\
      Stackoverflow & [40,000, 200,000] & [32, 32, 2] & 32 & 0.002 \\
      Taobao & [16,000, 80,000] & [32, 16, 4] & 32 & 0.002 \\
      USearthquake & [40,000, 200,000] & [32, 16, 4] & 32 & 0.002 \\
    \midrule
      Synthetic & [20,000, 100,000] & [32, 64, 3] & 32 & 0.002 \\
    \bottomrule
    \end{tabular}
    \vskip -0.1in
    \end{footnotesize}
\end{table}

\subsubsection{FullyNN Configurations}
\cref{tab:FullyNN_ete_hyp} shows hyperparameter settings for FullyNN. The three numbers in column ``MS" share the same meaning as those in \acrshort*{model}.
\begin{table}[!ht]
    \centering
    \caption{Hyperparameter settings for FullyNN.}
    \label{tab:FullyNN_ete_hyp}
    \begin{footnotesize}
    \begin{tabular}{ccccc}
    \toprule
      Datasets & Steps & MS & BS & LR \\
    \midrule
      Retweet & [80,000, 400,000] & [32, 16, 4] & 32 & 0.002 \\
      Stackoverflow & [40,000, 200,000] & [32, 32, 2] & 32 & 0.002 \\
      Taobao & [16,000, 80,000] & [32, 16, 4] & 32 & 0.002 \\
      USearthquake & [40,000, 200,000] & [32, 16, 4] & 32 & 0.002 \\
    \midrule
      Synthetic & [20,000, 100,000] & [32, 64, 3] & 32 & 0.002 \\
    \bottomrule
    \end{tabular}
    \vskip -0.1in
    \end{footnotesize}
\end{table}

\subsubsection{THP Configurations}
\cref{tab:thp_hyp} shows all hyperparameter settings for THP. The six values of ``MS" are the number of dimensions of the Transformer input vectors, the number of dimensions of the hidden outputs from an RNN which is on top of the Transformer encoder, the number of dimensions of the vectors used by self-attentions (\(q\), \(k\), and \(v\)), the number of Transformer layers, and heads.

\begin{table}[!ht]
    \centering
    \caption{Hyperparameter settings for THP.}
    \label{tab:thp_hyp}
    \begin{footnotesize}
    \begin{tabular}{ccccc}
    \toprule
      Datasets & Steps & MS & BS & LR \\
    \midrule
      Retweet & [80,000, 400,000] & [16, 16, 32, 8, 3, 3] & 32 & 0.002 \\
      Stackoverflow & [40,000, 200,000] & [16, 16, 32, 8, 3, 3] & 32 & 0.002 \\
      Taobao & [16,000, 80,000] & [16, 16, 32, 8, 3, 3] & 32 & 0.002 \\
      USearthquake & [40,000, 200,000] & [16, 16, 32, 8, 3, 3] & 32 & 0.002 \\
      \midrule
      Synthetic & [20,000, 100,000] & \makecell{[16, 32, 64, 16, 3, 4]} & 32 & 0.002 \\
    \bottomrule
    \end{tabular}
    \vskip -0.1in
    \end{footnotesize}
\end{table}

\subsubsection{SAHP Configurations}
The hyperparameter settings for SAHP are available in \cref{tab:sahp_hyp}. The first six values of ``MS" share the same meaning as those in THP while the last is the dropout rate.

\begin{table}[!ht]
    \centering
    \caption{Hyperparameter settings for SAHP.}
    \label{tab:sahp_hyp}
    \begin{footnotesize}
    \begin{tabular}{ccccc}
    \toprule
      Datasets & Steps & MS & BS & LR \\
    \midrule
      Retweet & [80,000, 400,000] & [16, 16, 32, 8, 3, 3, 0.1] & 32 & 0.002 \\
      Stackoverflow & [40,000, 200,000] & [16, 16, 32, 8, 3, 3, 0.1] & 32 & 0.002 \\
      Taobao & [16,000, 80,000] & [16, 16, 32, 8, 3, 3, 0.1] & 32 & 0.002 \\
      USearthquake & [40,000, 200,000] & [16, 16, 32, 8, 3, 3, 0.1] & 32 & 0.002 \\
      \midrule
      Synthetic & [20,000, 100,000] & \makecell{[16, 32, 64, 16, 3, 4, 0.1]} & 32 & 0.002 \\
    \bottomrule
    \end{tabular}
    \vskip -0.1in
    \end{footnotesize}
\end{table}

\subsubsection{AttNHP Configurations}
The hyperparameter settings for AttNHP are available in \cref{tab:attnhp_hyp}. The first six values of ``MS" share the same meaning as those in THP while the last is the dropout rate.

\begin{table}[!ht]
    \centering
    \caption{Hyperparameter settings for SAHP.}
    \label{tab:attnhp_hyp}
    \begin{footnotesize}
    \begin{tabular}{ccccc}
    \toprule
      Datasets & Steps & MS & BS & LR \\
    \midrule
      Retweet & [80,000, 400,000] & [16, 16, 64, 8, 3, 3, 0.0] & 32 & 0.002 \\
      Stackoverflow & [40,000, 200,000] & [16, 16, 64, 8, 3, 3, 0.0] & 4 & 0.002 \\
      Taobao & [16,000, 80,000] & [16, 16, 64, 8, 3, 3, 0.0] & 32 & 0.002 \\
      USearthquake & [40,000, 200,000] & [16, 16, 64, 8, 3, 3, 0.0] & 32 & 0.002 \\
      \midrule
      Synthetic & [20,000, 100,000] & \makecell{[16, 16, 64, 8, 3, 3, 0.0]} & 32 & 0.002 \\
    \bottomrule
    \end{tabular}
    \vskip -0.1in
    \end{footnotesize}
\end{table}

\subsubsection{Marked-LNM Configurations}
The hyperparameter settings for Marked-LNM are presented in \cref{tab:lnm_hyp}. The three values of ``MS" are the number of the dimensions of LSTM, the number of the dimensions of mark embedding, and the number of Gaussian distributions, respectively.

\begin{table}[!ht]
    \centering
    \caption{Hyperparameter settings for Marked-LNM.}
    \label{tab:lnm_hyp}
    \begin{footnotesize}
    \begin{tabular}{ccccc}
    \toprule
      Datasets & Steps & MS & BS & LR \\
    \midrule
      Retweet & [80,000, 400,000] & \makecell{[32, 32, 16]} & 32 & 0.002 \\
      Stackoverflow & [40,000, 200,000] & \makecell{[32, 32, 16]} & 32 & 0.002 \\
      Taobao & [16,000, 80,000] & \makecell{[32, 32, 16]} & 32 & 0.002 \\
      USearthquake & [40,000, 200,000] & \makecell{[32, 32, 16]} & 32 & 0.002 \\
      \midrule
      Synthetic & [20,000, 100,000] & \makecell{[32, 32, 16]} & 32 & 0.002 \\
    \bottomrule
    \end{tabular}
    \vskip -0.1in
    \end{footnotesize}
\end{table}

\section{Additional Experiment Results}
\label{app:additional}
\mysubsection[modeling joint distribution]{Performance of \acrshort*{model} for modeling \(p^*(m, t)\)}
\label{app:part1}
For a better and integration-free solution, \acrshort*{model} models the improper integration of \(p^*(m,t)\). The advantage has been verified by the experiment results reported in \cref{tab:IFIB_real_world_mae_e}. \acrshort*{model} models \(p^*(m,t)\) at the same time while modeling the improper integration of \(p^*(m,t)\). Compared to other existing \acrshort*{mtpp} models, the performance of \acrshort*{model} in modeling \(p^*(m,t)\) is evaluated and reported in \cref{tab:nll}. The evaluation metric is NLL loss, the average of the \(-\log{p}^*(m, t)\) at the observed events. The lower NLL loss indicates a better performance. We can observe that \acrshort*{model} shows a competent performance.
\begin{table*}[!ht]
    \caption{Accuracy of $p^*(m,t)$ measured by NLL loss on real-world datasets. Lower is better.}
   \vskip 0.15in
   \centering
   \begin{small}
   \setlength{\tabcolsep}{2pt}
   \begin{tabular}{lcccccc}
       \toprule
                                  & \acrshort*{model} (Ours) & FullyNN & SAHP & THP & AttNHP & Marked-LNM \\
       \midrule
       Retweet                    & {6.3225\tiny{$\pm$0.0007}} & 6.6437\tiny{$\pm$0.0380} & {6.1935\tiny{$\pm$0.0184}} & 10.379\tiny{$\pm$0.5349} & \textbf{\meanstd{6.0084}{0.0086}} & 6.5292\tiny{$\pm$0.0064} \\
       Stackoverflow              & \textbf{2.0540\tiny{$\pm$0.0029}} & 3.6984\tiny{$\pm$0.0022} & {2.0713\tiny{$\pm$0.0028}} & 2.5565\tiny{$\pm$0.0216} & 2.0811\tiny{$\pm$0.0054} & 2.0992\tiny{$\pm$0.0014} \\
       Taobao                     & {-0.7762\tiny{$\pm$0.0565}} & -0.0431\tiny{$\pm$0.0484} & \textbf{-1.2779\tiny{$\pm$0.0421}} & 140.91\tiny{$\pm$81.166} & -1.2190\tiny{$\pm$0.0763} & 1.2720\tiny{$\pm$0.1300} \\
       USearthquake               & \textbf{1.3278\tiny{$\pm$0.0533}} & 1.8664\tiny{$\pm$0.0649} & {1.3544\tiny{$\pm$0.0300}} & 2.0744\tiny{$\pm$0.3174} & 1.4120\tiny{$\pm$0.0499} & 1.8514\tiny{$\pm$0.0462} \\
       \bottomrule
   \end{tabular}
   \end{small}
   \label{tab:nll}
\vskip -0.1in
\end{table*}

\subsection{Evaluating Model Fidelity on Synthetic datasets}
\label{app:synthetic}

In this section, we report the full result of model fidelity test on synthetic datasets involving \acrshort*{model} and other baselines. The \acrshort*{model} consistently learns more accurate \(p^*(m, t)\) than other baselines as supported by the lower \(L^1\) distnace and higher Spearman coefficient. These findings suggest that predictions based on \acrshort*{model} should be more reliable and accurate.

\begin{table*}[!ht]
    \centering
    \caption{Model fidelity test performance on synthetic datasets; higher Spearman, lower \(L^1\) and relative NLL loss are better; the bold and underline indicate the best and the second-best values, respectively.}
    \label{table:synthetic}
    \begin{footnotesize}
    \begin{tabular}{llccccc}
        \toprule
         & & Hawkes\_1 & Hawkes\_2 & Poisson & Self-correct & Stationary Renewal \\
        \midrule
    \multirow{5}{*}{\rotatebox[origin=c]{90}{Spearman}} & \acrshort*{model} (Ours) & \textbf{1.0000\tiny{\(\pm\)0.0000}} & \textbf{0.9999\tiny{\(\pm\)0.0000}} & \textbf{1.0000\tiny{\(\pm\)0.0000}} & \textbf{0.9551\tiny{\(\pm\)0.0009}} & \textbf{0.9999\tiny{\(\pm\)0.0000}} \\
                              & FullyNN & 0.9952\tiny{\(\pm\)0.0004} & 0.9963\tiny{\(\pm\)0.0002} & 0.9722\tiny{\(\pm\)0.0018} & 0.9477\tiny{\(\pm\)0.0001} & \underline{0.9998\tiny{\(\pm\)0.0000}} \\
                              & SAHP & \underline{0.9959\tiny{\(\pm\)0.0047}} & 0.9862\tiny{\(\pm\)0.0000} & 0.9615\tiny{\(\pm\)0.0025} & \underline{0.9492\tiny{\(\pm\)0.0014}} & 0.9990\tiny{\(\pm\)0.0007} \\
                              & THP & 0.9266\tiny{\(\pm\)0.0026} & 0.7366\tiny{\(\pm\)0.0005} & \textbf{1.0000\tiny{\(\pm\)0.0000}} & 0.6969\tiny{\(\pm\)0.0017} & 0.0413\tiny{\(\pm\)0.0024} \\
                              & AttNHP & - & - & - & - & - \\
                              & Marked-LNM & 0.9924\tiny{\(\pm\)0.0007} & 0.9971\tiny{\(\pm\)0.0001} & 0.9713\tiny{\(\pm\)0.0024} & 0.9491\tiny{\(\pm\)0.0005} & \textbf{0.9999\tiny{\(\pm\)0.0000}} \\
        \midrule
    \multirow{5}{*}{\rotatebox[origin=c]{90}{\(L^1\)}}  & \acrshort*{model} (Ours) & \textbf{0.1480\tiny{\(\pm\)0.0085}} & \textbf{0.3105\tiny{\(\pm\)0.0432}} & \textbf{0.0133\tiny{\(\pm\)0.0091}} & \textbf{0.5163\tiny{\(\pm\)0.0290}} & \underline{0.0654\tiny{\(\pm\)0.0018}} \\
                              & FullyNN & \underline{0.6235\tiny{\(\pm\)0.0227}} & 3.1048\tiny{\(\pm\)0.0763} & 0.2973\tiny{\(\pm\)0.0098} & {1.1889\tiny{\(\pm\)0.0244}} & 0.0710\tiny{\(\pm\)0.0099} \\
                              & SAHP & 1.0245\tiny{\(\pm\)0.2967} & 4.7867\tiny{\(\pm\)0.2735} & 0.6893\tiny{\(\pm\)0.0238} & 1.3363\tiny{\(\pm\)0.0196} & 0.4872\tiny{\(\pm\)0.1833} \\
                              & THP & 12.003\tiny{\(\pm\)0.2069} & 25.500\tiny{\(\pm\)0.3642} & \underline{0.0203\tiny{\(\pm\)0.0067}} & 10.656\tiny{\(\pm\)0.0965} & 9.9230\tiny{\(\pm\)0.0451} \\
                              & AttNHP & - & - & - & - & - \\
                              & Marked-LNM & 0.6994\tiny{\(\pm\)0.0117} & 2.6446\tiny{\(\pm\)0.0633} & 0.3620\tiny{\(\pm\)0.0044} & \underline{0.7406\tiny{\(\pm\)0.0168}} & \textbf{0.0402\tiny{\(\pm\)0.0001}} \\
        \midrule
    \multirow{5}{*}{\rotatebox[origin=c]{90}{Relative NLL}} & \acrshort*{model} (Ours) & \textbf{0.0000\tiny{\(\pm\)0.0000}} & \textbf{0.0001\tiny{\(\pm\)0.0000}} & \textbf{0.0000\tiny{\(\pm\)0.0000}} & \textbf{0.0007\tiny{\(\pm\)0.0003}} & \textbf{0.0000\tiny{\(\pm\)0.0000}} \\
                              & FullyNN &\underline{0.0003\tiny{\(\pm\)0.0000}} & \underline{0.0008\tiny{\(\pm\)0.0001}} &0.0002\tiny{\(\pm\)0.0000} & \underline{0.0015\tiny{\(\pm\)0.0001}} & \textbf{0.0000\tiny{\(\pm\)0.0000}} \\
                              & SAHP & 0.0086\tiny{\(\pm\)0.0017} & 0.0312\tiny{\(\pm\)0.0193} & 0.0092\tiny{\(\pm\)0.0002} & 0.0072\tiny{\(\pm\)0.0009} & \underline{0.0034\tiny{\(\pm\)0.0010}} \\
                              & THP & 0.2137\tiny{\(\pm\)0.0001}& 0.6663\tiny{\(\pm\)0.0029} & \textbf{0.0000\tiny{\(\pm\)0.0000}} & 0.1262\tiny{\(\pm\)0.0004} & 0.0771\tiny{\(\pm\)0.0000} \\
                              & AttNHP & 0.8202\tiny{\(\pm\)0.0053}& 0.0387\tiny{\(\pm\)0.0144} & 0.2631\tiny{\(\pm\)0.0009} & 0.0820\tiny{\(\pm\)0.0003} & {0.3065\tiny{\(\pm\)0.0007}} \\
                              & Marked-LNM & 0.0004\tiny{\(\pm\)0.0000}& 0.0010\tiny{\(\pm\)0.0000} & 0.0006\tiny{\(\pm\)0.0000} & 0.0018\tiny{\(\pm\)0.0001} & \textbf{0.0000\tiny{\(\pm\)0.0000}} \\
        \bottomrule
    \end{tabular}
    \end{footnotesize}
    \vskip -0.1in
    \end{table*}

\subsection{Comparing IFNMTPP with Marked-LNM and thresholding}
\label{app:with_marked-lnm}
Among all baselines, only Marked-LNM follows the mark-time modeling paradigm and is suitable with thresholding. We therefore compare the mark prediction accuracy of our method and Marked-LNM under thresholding. The results, summarized in the table below, demonstrate that our method outperforms Marked-LNM in mark prediction. Note that we do not report time prediction results for this comparison, as the time prediction is unaffected by the method used for predicting marks. Nonetheless, as shown in Table 6, our method also achieves strong performance in time prediction.

\begin{table}[!ht]
\centering
\caption{Comparison of \acrshort*{model} with Lognormmix + thresholding on four data sets, measured by macro-F1/micro-F1. The bold are the best values.}
\label{tab:comparison}
\resizebox{\linewidth}{!}{
\begin{tabular}{llcccc}
\toprule
 & & Retweet & SO & Taobao & USearthquake \\
\midrule
\multirow{3}{*}{ours} & $M$ & \textbf{0.4750$\pm$0.0033} / \textbf{0.4394$\pm$0.0093} & \textbf{0.1776$\pm$0.0030} / \textbf{0.6376$\pm$0.0026} & \textbf{0.4190$\pm$0.0104} / 0.7499$\pm$0.0151 & \textbf{0.1382$\pm$0.0071} / 0.3189$\pm$0.0125 \\
 & $M_r$ & \textbf{0.2010$\pm$0.0082} / \textbf{0.2010$\pm$0.0082} & \textbf{0.1476$\pm$0.0041} / \textbf{0.4530$\pm$0.0026} & \textbf{0.3987$\pm$0.0108} / 0.7558$\pm$0.0185 & 0.0339$\pm$0.0051 / 0.1111$\pm$0.0098 \\
 & $M_f$ & \textbf{0.6120$\pm$0.0013} / \textbf{0.9612$\pm$0.0021} & \textbf{0.2795$\pm$0.0014} / \textbf{0.8974$\pm$0.0042} & 0.7441$\pm$0.0060 / 0.7441$\pm$0.0060 & \textbf{0.2773$\pm$0.0215} / \textbf{0.9181$\pm$0.0102} \\
\midrule
\multirow{3}{*}{\makecell{Makred-LNM +\\ thresholding}} & $M$ & 0.4228$\pm$0.0014 / 0.3876$\pm$0.0093 & 0.1121$\pm$0.0007 / 0.4469$\pm$0.0124 & 0.1558$\pm$0.0623 / \textbf{0.7945$\pm$0.0060} & 0.1198$\pm$0.0078 / \textbf{0.3438$\pm$0.0035} \\
 & $M_r$ & 0.1730$\pm$0.0033 / 0.1730$\pm$0.0033 & 0.1004$\pm$0.0004 / 0.3527$\pm$0.0065 & 0.1181$\pm$0.0653 / \textbf{0.8318$\pm$0.0019} & \textbf{0.0393$\pm$0.0004} / \textbf{0.1740$\pm$0.0074} \\
 & $M_f$ & 0.5477$\pm$0.0004 / 0.8687$\pm$0.0005 & 0.1519$\pm$0.0017 / 0.5662$\pm$0.0209 & \textbf{0.7589$\pm$0.0133} / \textbf{0.7589$\pm$0.0133} & 0.2271$\pm$0.0169 / 0.6810$\pm$0.0427 \\
\bottomrule
\end{tabular}
}
\end{table}

\end{document}